\documentclass[letterpaper, journal]{IEEEtran}
\usepackage{amsmath,amsfonts}
\usepackage{algorithmic}
\usepackage{array}
\usepackage[caption=false,font=normalsize,labelfont=sf,textfont=sf]{subfig}
\usepackage{textcomp}
\usepackage{stfloats}
\usepackage{url}
\usepackage{verbatim}
\usepackage{graphicx}
\usepackage{booktabs}
\usepackage{multirow}
\usepackage{xcolor}
\usepackage{makecell}

\usepackage{threeparttable}
\usepackage{makecell}
\usepackage{booktabs}
\usepackage{multirow}

\usepackage{orcidlink}

\usepackage[table]{xcolor} 
\usepackage{booktabs}
\definecolor{lightpurple}{RGB}{230, 230, 255}
\usepackage{tabularx}

\def\BibTeX{{\rm B\kern-.05em{\sc i\kern-.025em b}\kern-.08em
    T\kern-.1667em\lower.7ex\hbox{E}\kern-.125emX}}
\usepackage{balance}
\begin{document}

\title{
WUTDet: A 100K-Scale Ship Detection Dataset and Benchmarks with Dense Small Objects
}

\author{Junxiong Liang$^{\dagger}$ \hspace{-2.0mm}$^{~\orcidlink{0009-0008-4218-7332}}$, Mengwei Bao$^{\dagger}$ \hspace{-2.0mm}$^{~\orcidlink{0009-0008-5658-8962}}$, Tianxiang Wang \hspace{-2.0mm}$^{~\orcidlink{0009-0008-4045-4521}}$, Xinggang Wang \hspace{-2.0mm}$^{~\orcidlink{0000-0001-6732-7823}}$, \IEEEmembership{Senior Member, IEEE}, \\An-An Liu \hspace{-2.0mm}$^{~\orcidlink{0000-0001-5755-9145}}$, \IEEEmembership{Senior Member, IEEE}, and Ryan Wen Liu$^*$ \hspace{-2.0mm}$^{~\orcidlink{0000-0002-1591-5583}}$, \IEEEmembership{Member, IEEE}


\thanks{This work was supported in part by the National Natural Science Foundation of China under Grant 52422111 and Grant 52271365, and in part by the Excellent Youth Foundation of Hubei Scientific Committee under Grant 2024AFA042. (Corresponding author: Ryan Wen Liu.)}
    
\thanks{Junxiong Liang, Mengwei Bao, Tianxiang wang and Ryan Wen Liu are with School of Navigation, Wuhan University of Technology, Wuhan 430063, China, with Sanya Science and Education Innovation Park, Wuhan University of Technology, Sanya 572000, China, and also with the State Key Labora tory of Maritime Technology and Safety, Wuhan 430063, China (e-mail: jxliang@whut.edu.cn; mengweibao@whut.edu.cn; tianxiangwang@whut.edu.cn; wenliu@whut.edu.cn).}

\thanks{Xinggang Wang is with the School of Electronic Information and Communications, Huazhong University of Science and Technology, Wuhan 430074, China (e-mail: xgwang@hust.edu.cn).}

\thanks{An-An Liu is with the School of Electrical and Information Engineering, Tianjin University, Tianjin 300072, China (e-mail: anan0422@gmail.com).}

\thanks{{$^{\dagger}$ Equal contribution.}}

\thanks{{$^*$ Corresponding author.}}

}


\maketitle

\begin{abstract}

    Ship detection for navigation is a fundamental perception task in intelligent waterway transportation systems. However, existing public ship detection datasets remain limited in terms of scale, the proportion of small-object instances, and scene diversity, which hinders the systematic evaluation and generalization study of detection algorithms in complex maritime environments. To this end, we construct WUTDet, a large-scale ship detection dataset. WUTDet contains 100,576 images and 381,378 annotated ship instances, covering diverse operational scenarios such as ports, anchorages, navigation, and berthing, as well as various imaging conditions including fog, glare, low-lightness, and rain, thereby exhibiting substantial diversity and challenge. Based on WUTDet, we systematically evaluate 20 baseline models from three mainstream detection architectures, namely CNN, Transformer, and Mamba. Experimental results show that the Transformer architecture achieves superior overall detection accuracy ($AP$) and small-object detection performance ($AP_s$), demonstrating stronger adaptability to complex maritime scenes; the CNN architecture maintains an advantage in inference efficiency, making it more suitable for real-time applications; and the Mamba architecture achieves a favorable balance between detection accuracy and computational efficiency. Furthermore, we construct a unified cross-dataset test set, Ship-GEN, to evaluate model generalization. Results on Ship-GEN show that models trained on WUTDet exhibit stronger generalization under different data distributions. These findings demonstrate that WUTDet provides effective data support for the research, evaluation, and generalization analysis of ship detection algorithms in complex maritime scenarios. The dataset is publicly available at: https://github.com/MAPGroup/WUTDet.
    
\end{abstract}


\section{Introduction} \label{sec: introduction}

    \IEEEPARstart{W}{ith} the rapid development of intelligent ships and unmanned navigation technologies, vision-based environmental perception has become a critical component for achieving autonomous navigation and safe collision avoidance \cite{lin2022maritime}. As one of the core tasks in visual perception, object detection aims to automatically localize and recognize ships in complex maritime scenes, providing essential information for subsequent target tracking, behavior analysis, and trajectory planning \cite{rekavandi2025guide, wang2024mdd}. In real-world navigation environments, ship detection results directly affect the reliability of collision risk assessment and avoidance decisions, playing a vital role in improving navigation safety, operational efficiency, and the intelligence of waterway transportation systems \cite{zhang2025deep, wang2024aodemar}. However, maritime scenes are often characterized by complex backgrounds, drastic lightness variations, changing weather conditions, and large-scale differences in target sizes, which pose significant challenges to the robustness and generalization capability of ship detection algorithms.

    \begin{figure}
        \centering
        \includegraphics[width=1.0\linewidth]{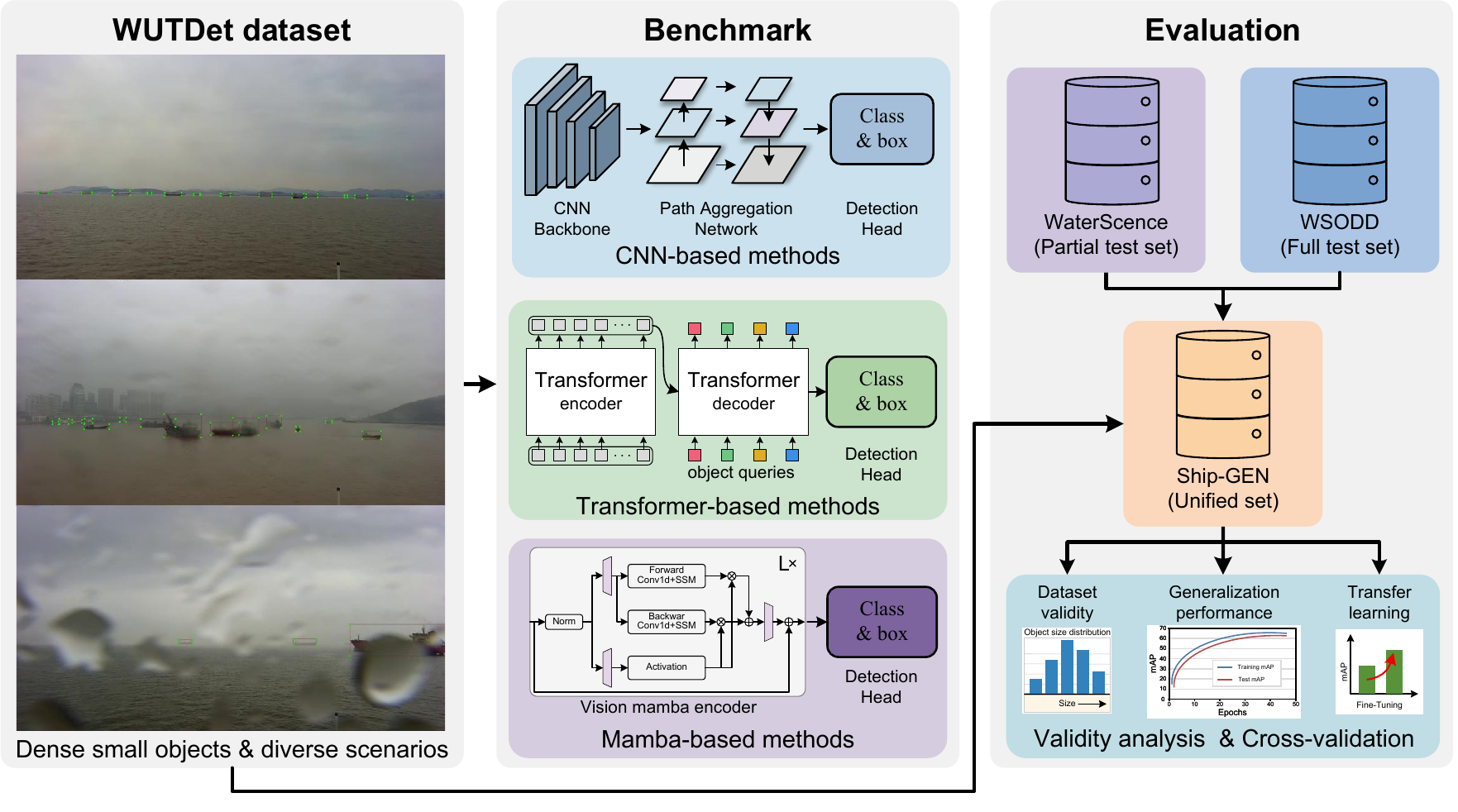}
        \caption{Overview of the proposed framework, including the construction of the large-scale ship detection dataset (WUTDet), the systematic evaluation of detection models with different architectures, and the construction and validation of the cross-dataset test set (Ship-GEN). WUTDet is characterized by dense small objects and diverse scenarios, while Ship-GEN is used to validate the effectiveness of WUTDet and the generalization capability of the models.}
        \label{fig: overview}
    \end{figure}

    In recent years, deep learning has driven the rapid evolution of object detection methods. Early convolutional neural network (CNN)–based detectors, such as Faster R-CNN \cite{ren2015faster}, YOLO \cite{redmon2018yolov3, bochkovskiy2020yolov4}, and SSD \cite{liu2016ssd}, achieved remarkable success in general object detection tasks by extracting local features through multi-layer convolutional structures. Subsequently, the Transformer \cite{vaswani2017attention} architecture was introduced into the vision domain \cite{dosovitskiy2020image}. Its self-attention mechanism enables the modeling of long-range dependencies, endowing models with stronger global feature representation capability. Representative methods include DETR \cite{carion2020end} and its improved variants \cite{zhu2020deformable, zhao2024detrs}. More recently, the Mamba \cite{gu2024mamba} architecture based on state space models (SSM) has demonstrated advantages in long-sequence modeling and efficient inference, providing a new technical paradigm for object detection \cite{zhu2024vision}.
    CNN-, Transformer-, and Mamba-based architectures differ substantially in feature representation and computational mechanisms. Their respective performance in ship detection tasks thus warrants systematic evaluation and comparative analysis.

    Although deep learning–based detection models have achieved continuous advances, their performance remains highly dependent on large-scale and high-quality training datasets. However, existing public ship detection datasets \cite{shao2018seaships, zhou2021image, yao2024waterscenes} still exhibit significant limitations in terms of scale and diversity. On the one hand, some datasets contain a limited number of samples, which is insufficient to support the training of large models and reliable performance evaluation. On the other hand, with respect to target scale distribution, small-scale ship targets are generally underrepresented, failing to reflect real navigation scenarios in which distant vessels dominate. In addition, some datasets provide insufficient coverage of ship operational scenarios and imaging conditions, which restricts the generalization capability of detection models in complex maritime environments. Therefore, constructing a ship detection dataset with larger scale, richer scene diversity, and more fine-grained annotations is of great importance for systematically evaluating the robustness and generalization ability of detection models under complex maritime conditions.

    Based on the above analysis, this paper constructs a large-scale ship detection dataset (WUTDet) with diverse scenarios and target scales. Based on WUTDet, we conduct a systematic evaluation of detection models with different architectures and further establish a cross-dataset test set (Ship-GEN) to validate the effectiveness of WUTDet and the generalization capability of the models. The overall framework is illustrated in Fig. \ref{fig: overview}. Through joint analysis at both the data and algorithm levels, this work reveals the performance differences and applicability of mainstream detection models in ship detection tasks, and provides quantitative references for subsequent model design and methodological improvement. The main contributions of this paper are summarized as follows:

    \begin{itemize}
        \item We construct a large-scale, high-resolution ship detection dataset (WUTDet), which contains 100,576 images and 381,378 ship instances. The dataset provides fine-grained annotations of ship targets across diverse operational scenarios, imaging conditions, and target scales. In addition, we perform a comprehensive statistical analysis from multiple perspectives, including target-scale distribution and scene composition, thereby further revealing the main challenges involved in ship detection tasks.

        \item Based on the constructed dataset, we conduct comparative experiments and performance analysis on 20 baseline object detection models, covering representative detection methods based on CNN, Transformer, and Mamba architectures. Through in-depth analysis of the experimental results, we summarize the performance differences of these models in ship detection tasks, together with their respective strengths and limitations, thereby providing quantitative evidence and benchmark references for subsequent research and methodological improvements.

        \item To validate the effectiveness of WUTDet and systematically evaluate model generalization, we construct a unified n cross-dataset test set (Ship-GEN). Models are trained on different datasets and then evaluated on Ship-GEN under a unified testing protocol, thereby assessing their generalization capability in cross-dataset scenarios.
    \end{itemize}

    The remainder of this paper is organized as follows. Section \ref{sec: related work} reviews related work on object detection datasets and detection algorithms. Section \ref{sec: dataset} presents the constructed ship detection dataset (WUTDet), including data collection, data processing and annotation, and dataset Statistics. Section \ref{sec: method} describes the 20 baseline detection algorithms adopted in the experiments. Section \ref{sec: experiments} reports the experimental results of these baseline methods on the WUTDet dataset and further presents their performance on the constructed Ship-GEN dataset. Finally, Section \ref{sec: conclusion} concludes the paper.

    \begin{table*}[htbp]
        \centering
        \renewcommand{\arraystretch}{1.5}
        \setlength{\tabcolsep}{2pt}
        \caption{Overview of Ship Detection–Related Datasets. This table provides a statistical comparison of different datasets in terms of release year, number of images, number of annotated objects, image resolution, annotation categories, and the quantity and proportion of targets at different scales. Here, "$\dagger$" indicates that the corresponding dataset is unavailable, and “random” denotes that the dataset contains images with more than two different resolutions, with most samples having a resolution lower than 1920$\times$1080. Bold indicate the highest values in the corresponding columns.}
        \label{tab: dataset comparison}
        \begin{threeparttable}
            \begin{tabular*}{\textwidth}{@{\extracolsep{\fill}} l c c c c c c c c c c c}
            \toprule
            \multirow{2}{*}{Dataset} & \multirow{2}{*}{Year} & \multirow{2}{*}{Image/Frame} & \multirow{2}{*}{Object} & \multirow{2}{*}{Size}                                                                   & \multirow{2}{*}{Category}  & \multicolumn{6}{c}{Object Size}                                                                                                                                       \\
                                                                                                                                                                                                                                             \cline{7-12}
                                                                    &                       &                        &                               &                                                          &                            & \multicolumn{2}{c}{$0\!\sim\!32^2$ (Small)} & \multicolumn{2}{c}{$32^2\!\sim\!96^2$ (Medium)} & \multicolumn{2}{c}{$>96^2$ (Large)}                                   \\
            \midrule
            SMD \cite{prasad2017video}                              & 2017                  & 20,367                 & 157,998                       & 1920$\times$1080                                         & 10                         & 16,017            & 10.14\%                 & 77,263            & 48.90\%                     & 6,4718            & 40.96\%                                           \\
            Seaships$^{\dagger}$ \cite{shao2018seaships}            & 2018                  & 31,455                 & 40,077                        & 1920$\times$1080                                         & 6                          & -                 & -                       & -                 & -                           & -                 & -                                                 \\
            SeaShips(7000) \cite{shao2018seaships}                  & 2018                  & 7,000                  & 9,221                         & 1920$\times$1080                                         & 6                          & 15                & 0.16\%                  & 784               & 8.50\%                      & 8,422             & 91.33\%                                           \\
            Mcships$^{\dagger}$ \cite{zheng2020mcships}             & 2020                  & 14,709                 & 26,529                        & random                                                   & 13                         & -                 & -                       & -                 & -                           & -                 & -                                                 \\
            Mcships-lite \cite{zheng2020mcships}                    & 2020                  & 7,996                  & 11,331                        & random                                                   & 2                          & 949               & 8.38\%                  & 2,038             & 17.99\%                     & 8,344             & 73.64\%                                           \\ 
            WSODD \cite{zhou2021image}                              & 2021                  & 7,467                  & 21.911                        & random                                                   & 14                         & 2,984             & 13.62\%                 & 9443              & 43.10\%                     & 9,484             & 43.28\%                                           \\
            SimuShips \cite{raza2022simuships}                      & 2022                  & 9,471                  & 22503                         & 1920$\times$1080                                         & 2                          & 3006              & 13.36\%                 & 4138              & 18.39\%                     & 15359             & 68.25\%                                           \\
            FVessel \cite{guo2023asynchronous}                      & 2023                  & 11,210                 & 28,213                        & 2560$\times$1440                                         & 1                          & 2,472             & 8.76\%                  & 14,744            & \textbf{52.26\%}            & 10,997            & 38.98\%                                           \\
            WaterScenes \cite{yao2024waterscenes}                   & 2023                  & 54,120                 & 202,807                       & 1920$\times$1080                                         & 7                          & 53,348            & 26.30\%                 & 91,287            & 45.01\%                     & 58,172            & 28.68\%                                           \\
            MVDD13 \cite{wang2024marine}                            & 2024                  & 35,474                 & 40,869                        & random                                                   & 13                         & 189               & 0.46\%                  & 2018              & 4.94\%                      & 38,662            & \textbf{94.60\%}                                  \\
            \midrule
            WUTDet(Ours)                                           & 2026                  & 100,576                & 381,378                       & \makecell{1920$\times$1080 \\ 2560$\times$1440}          & 1                          & \textbf{132,882}   & \textbf{34.84\%}        & \textbf{163,642}  & 42.91\%                    & \textbf{84,854}    & 22.25\%                                           \\
            \bottomrule
            \end{tabular*}
            
            \begin{tablenotes}
            \footnotesize
            \item Note: All ship instances in WUTDet are annotated as a single category, "ship." This design is based on two primary considerations. First, fine-grained ship categorization is not a critical factor for collision avoidance decision-making in autonomous maritime navigation. Second, the dataset contains a high prevalence of small-scale objects (area $\le 32^2$ pixels) that lack sufficient visual semantic information for reliable fine-grained annotation. 
            \end{tablenotes}
            
        \end{threeparttable}
    \end{table*}

\section{Related Work} \label{sec: related work}
    
    This section provides a review and analysis of existing related work from the perspectives of object detection datasets and object detection methods.
    
    \subsection{Object Detection Dataset}
        \textbf{General Object Detection Datasets.} General-purpose object detection datasets provide unified benchmarks for training and evaluation, serving as fundamental support for the development of detection algorithms in terms of model architecture design and performance improvement. PASCAL VOC \cite{everingham2010pascal} is one of the earliest representative datasets, containing 20 common object categories and establishing a standardized evaluation protocol for object detection. It laid an important foundation for the development of two-stage detection frameworks. Owing to its moderate scale and high annotation quality, this dataset remained a widely adopted benchmark for performance comparison over an extended period. Building upon this, ImageNet \cite{deng2009imagenet} extended object detection annotations through the ILSVRC \cite{russakovsky2015imagenet} detection subset, whose large-scale labeled samples significantly promoted the application of deep convolutional neural networks in visual tasks and drove detection models toward deeper and more complex architectures. Subsequently, MS COCO \cite{lin2014microsoft} further increased the scale and complexity of detection datasets, covering 80 object categories. Its distinctive characteristics include a more diverse object scale distribution, a higher proportion of small objects, and an emphasis on detecting densely distributed multiple objects under complex background conditions, making it one of the most widely used benchmark datasets in current object detection research.To further expand category diversity and semantic coverage, Open Images \cite{kuznetsova2020open} continuously increased both the number of samples and object categories, containing approximately 6,000 object instances with hierarchical annotation structures, thereby providing important support for large-scale model training and generalization capability evaluation. Meanwhile, Objects365 \cite{shao2019objects365} aims to construct an ultra-large-scale object detection dataset, including 365 object categories and over 10 million bounding box annotations. It compensates for the limitations of traditional datasets in terms of category scale and scene coverage, offering a more sufficient data foundation for training large-scale detection models.
        
        Despite their significant value in algorithm design and performance evaluation, these general-purpose datasets are mainly collected from terrestrial and urban environments, with object categories dominated by pedestrians, vehicles, and everyday objects. Their imaging conditions and scene structures differ substantially from real maritime environments. Such datasets usually lack typical maritime characteristics, such as strong sea-surface reflections, complex meteorological conditions, and long-range small-scale ships. Consequently, relying solely on general-purpose datasets is insufficient for comprehensively evaluating the robustness of detection models in complex maritime scenarios. Therefore, it is necessary to construct specialized object detection datasets for maritime environments as an important complement.
    
        \textbf{Ship Object Detection Datasets.} Compared with general-purpose object detection datasets, specialized datasets constructed for maritime scenarios can more accurately reflect the imaging characteristics of water surfaces and the distribution patterns of ship targets. From a temporal perspective, maritime datasets have evolved from basic environmental perception to multi-task integration, and further toward large-scale fine-grained recognition. In 2015, MODD \cite{kristan2015fast}, as a pioneering work, established a fundamental framework for water-surface obstacle detection using 12 waterborne video sequences. Although its targets were only coarsely divided into “large obstacles” and “small obstacles,” it laid an initial foundation for vision-based collision avoidance research. Subsequently, SMD \cite{prasad2017video}, released in 2017, significantly enriched the diversity of maritime scenarios by covering ten target categories from both shore-based and onboard viewpoints in Singapore waters, and introduced near-infrared imaging to support target perception under low-lighnetness conditions. In 2018, the release of SeaShips \cite{shao2018seaships} marked the maturity of large-scale, precisely annotated ship datasets. It collected 31,455 images from port surveillance systems and covered six common inland vessel categories. After 2020, datasets gradually evolved toward more complex and multi-dimensional tasks. McShips \cite{zheng2020mcships} provided fine-grained annotations of 13 military and civilian ship categories, posing greater challenges to detection models under conditions of minimal inter-class differences. WSODD \cite{zhou2021image} enhanced environmental diversity by covering ports, nearshore areas, and open waters, enabling better characterization of complex backgrounds and diverse water-surface conditions. SimuShips \cite{raza2022simuships} presents a simulation-based ship detection dataset containing 9,471 images with precise bounding-box annotations. By systematically incorporating diverse obstacle types, weather and illumination variations, occlusion, and scale changes, it provides a low-cost supplement to real-world maritime data. FVessel \cite{guo2023asynchronous} released in 2023 focused on improving vessel monitoring accuracy through asynchronous trajectory matching between AIS and video data. The latest WaterScenes \cite{yao2024waterscenes} further advances multi-task 4D radar–camera fusion perception for water-surface scenarios, whereas MVDD13 \cite{wang2024marine} establishes a large-scale, fine-grained visual benchmark for ship detection in USV applications.

        However, according to the statistics in Table \ref{tab: dataset comparison}, existing ship-specific datasets still exhibit significant limitations in supporting the training of advanced deep learning models (Transformer and Mamba architectures) and in enabling robust generalization evaluation under complex scenarios. First, in terms of data scale and availability, most public datasets remain at the order of tens of thousands of images. Moreover, due to copyright or privacy restrictions, some datasets are only partially released, such as SeaShips (7000) and McShips-lite, making it difficult to meet the data volume requirements of large-parameter models. Second, the distribution of target scales is generally highly imbalanced. Large objects account for 91.33\% and 94.60\% of the samples in SeaShips(7000) and MVDD13, respectively, whereas small objects constitute only 0.16\% and 0.46\%. In SMD, SimuShips, and FVessel, the proportion of small objects is also only 10.14\%, 13.36\%, and 8.76\%, respectively. This sample distribution, which overemphasizes large objects while neglecting small ones, fails to reflect real navigation scenarios where distant ships dominate, thereby constraining the detection performance of models on small ship. Furthermore, in terms of imaging resolution and environmental complexity, most datasets mainly adopt a single resolution of 1920$\times$1080, with insufficient coverage of higher resolutions such as 2560$\times$1440. To address these issues, we construct a large-scale ship detection dataset (WUTDet), aiming to systematically alleviate the problems of insufficient sample scale, biased target scale distribution, and limited imaging conditions in existing datasets. It provides a more challenging and application-oriented unified benchmark for ship detection algorithms in complex maritime environments.

        \begin{figure}[!t]
            \centering
            \includegraphics[width=1.0\linewidth]{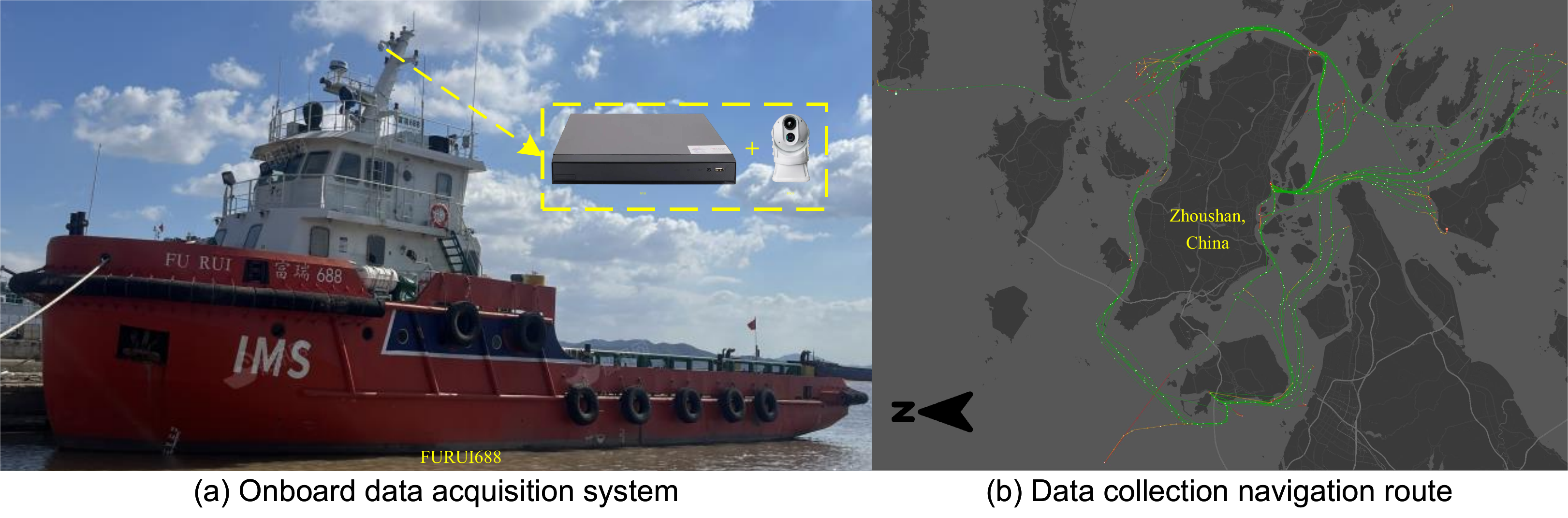}
            \caption{WUTDet Data Collection Setup. (a) Onboard data acquisition device used for image collection; (b) illustration of the ship navigation route over three months during data collection process.}
            \label{fig: data collection setup}
        \end{figure}

    \subsection{Object Detection Methods}
        \textbf{CNN-Based Methods.} CNN-based methods currently constitute the dominant paradigm for ship detection and can be broadly divided into two-stage and one-stage approaches. Two-stage methods, represented by Faster R-CNN \cite{ren2015faster} and Cascade R-CNN \cite{cai2018cascade}, generate region proposals followed by refined classification and regression, achieving superior localization accuracy. They often incorporate feature pyramid networks (FPN) or contextual modeling to alleviate challenges posed by large scale variations of ships and interference from complex sea backgrounds. In contrast, one-stage methods such as the YOLOs \cite{ge2021yolox, tian2025yolov12, lei2025yolov13} and SSD \cite{liu2016ssd} reformulate detection as a regression problem, greatly improving inference efficiency and meeting the real-time requirements of maritime surveillance. To accommodate the limited computational resources of shipborne embedded platforms, recent variants from YOLOv8 to YOLOv13 have continuously optimized lightweight backbone networks (C2f and C3K2 structures) and combined them with multi-scale feature representations and relational modeling, aiming to enhance robustness under complex sea conditions, particularly for detecting small vessels that are easily disturbed by waves. 

        \textbf{Transformer-Based Methods.} Transformer-based detection methods establish long-range dependencies through self-attention mechanisms and thus possess stronger global feature modeling capabilities. Since DETR \cite{carion2020end} reformulated object detection as a set prediction problem, research in this field has mainly focused on improving efficiency and achieving real-time performance. The first line of work, exemplified by Deformable DETR \cite{zhu2020deformable}, accelerates convergence through sparse sampling attention and significantly improves the detection performance of small, distant ships on the sea surface. The second line, represented by DAB-DETR \cite{liu2022dab} and DINO \cite{zhang2022dino}, enhances matching stability in complex backgrounds by refining object queries and introducing denoising training strategies. Furthermore, to meet practical deployment requirements, models such as RT-DETR \cite{zhao2024detrs} and D-FINE \cite{peng2024d} optimize bounding box regression and inference latency while preserving the advantages of end-to-end architectures. They demonstrate stronger feature selection and representation capability than traditional CNN-based methods when dealing with challenging scenarios such as strong sea-surface reflections and occlusions.

        \textbf{Mamba-Based Methods.} Mamba \cite{gu2024mamba}, as an emerging sequence modeling approach based on SSM, offers near-linear computational complexity, providing an efficient technical pathway for processing high-resolution maritime images. In ship detection tasks, architectures such as Vision Mamba \cite{zhu2024vision} and VMamba \cite{liu2024vmamba} leverage SSM to enhance visual representation learning, overcoming the limitation of Transformers whose computational cost grows quadratically with image resolution while maintaining efficient inference. Recently, Mamba architectures have begun to integrate with one-stage real-time detection frameworks (Mamba YOLO \cite{wang2025mamba}). By adopting SSM-based backbones, they reduce the computational burden of self-attention mechanisms and thereby enable low-latency object detection under real-time constraints. When applied to large-scale, high-resolution ship datasets (Ours WUTDet), such architectures demonstrate significant potential for balancing computational efficiency and detection robustness in complex maritime environments.

\section{Dataset} \label{sec: dataset}

    This paper aims to construct a large-scale real-world ship detection dataset (WUTDet) for maritime environment perception, collision avoidance, and safe navigation, and to conduct a evaluation of state-of-the-art object detection algorithms based on this dataset. As summarized in Table \ref{tab: dataset comparison}, WUTDet significantly exceeds existing ship detection datasets in terms of both the number of images and annotated object instances, and it contains a large proportion of small-scale ships at long distances, which better reflects the target distribution in real navigation scenarios. In this section, we describe the dataset construction process in detail and provide a statistical analysis of its composition and characteristics.
    
    \subsection{Data Collection}
    Visible-light cameras can capture high-resolution images of ship navigation areas and provide rich color and texture information. In addition, they offer advantages such as low power consumption, flexible deployment, and long-term stable operation. Therefore, the dataset constructed in this study uniformly adopts visible-light images as the data source. The data acquisition system consists of the Furui 688 vessel platform, the DN20 marine photoelectric evidence system, and a Hikvision network video recorder (NVR). The Furui 688 is a port supply vessel with a length of 35 m and a width of 8 m, serving as the data collection platform. The DN20 and the NVR are installed on the vessel to acquire and store video data in real time during navigation. The DN20 is equipped with a high-definition visible-light camera and an infrared thermal imager; in this study, only the video data captured from its visible-light channel are used. The NVR is responsible for the real-time storage of the video data collected by the DN20. Fig. \ref{fig: data collection setup} illustrates the overall data acquisition setup and navigation route. Specifically, Fig. \ref{fig: data collection setup} (a) shows the onboard data acquisition system, while Fig. \ref{fig: data collection setup} (b) presents the vessel’s navigation routes in the over a three-month data collection period. All images with a resolution of 1920$\times$1080 in the dataset are obtained from this acquisition system.

    \begin{figure*}[!t]
        \centering
        \includegraphics[width=1.0\linewidth]{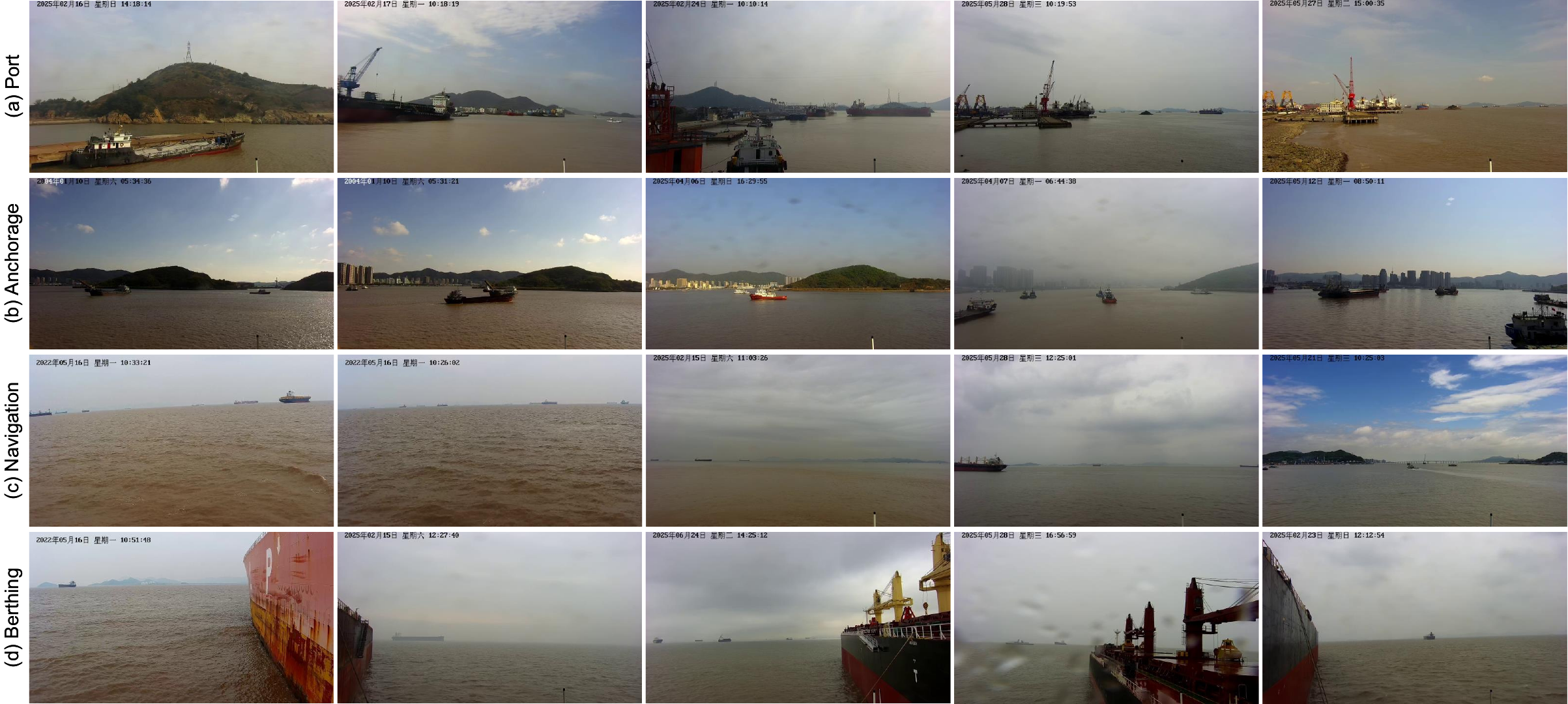}
        \caption{Image examples in WUTDet from four typical operational scenarios, namely port, anchorage, navigation, and berthing. These scenarios differ significantly in ship distribution, target scale, and occlusion level, posing diverse challenges for ship detection.}
        \label{fig: Samples in WUTDet}
    \end{figure*}

    \begin{table*}[!t]
        \centering
        \caption{Train/Validation/Test Split Statistics of WUTDet. Considering that ship type is of limited importance for collision avoidance in autonomous navigation and that small-scale objects do not provide sufficient visual semantic information for reliable fine-grained annotation, all ship instances in WUTDet are uniformly annotated as a single category.}
        \renewcommand{\arraystretch}{1.5}
        \setlength{\tabcolsep}{2pt}
        \label{tab: data split}
        \begin{tabular*}{\textwidth}{@{\extracolsep{\fill}} l c c c c c c }
        \hline
        Split          & 1920$\times$1080          & 2560$\times$1440          & Images          & Percentage          & Objects          & Category          \\
        \hline
        Train          & 65,070                    & 15,391                    & 80,461          & 80.0\%              & 305,264          & Ship              \\
        Validation     & 8,134                     & 1,924                     & 10,058          & 10.0\%              & 38,118           & Ship              \\
        Test           & 8,133                     & 1,924                     & 10,057          & 10.0\%              & 37,996           & Ship              \\
        \hline
        Total          & 81,337                    & 19,239                    & 100,576         & 100\%               & 381,378          & Ship              \\
        \hline
        \end{tabular*}
    \end{table*}

    To ensure the diversity and representativeness of the dataset, the onboard DN20 system conducted continuous video acquisition under a wide range of conditions, including different maritime scenarios (harbors, anchorages, berthing, and navigation states), different time periods (daytime and nighttime), varying lightness conditions (low-lightness and glare), and diverse weather conditions (fog and rain). In addition, for certain scenarios, stitched videos with a resolution of 5120$\times$1440 were further cropped to generate high-resolution images with a resolution of 2560$\times$1440. The higher image resolution provides richer target detail information, which is beneficial for training detection models with improved performance and stronger robustness. It should be noted that these stitched videos were also collected during actual vessel navigation. All video data were captured in the Zhoushan waters.

    \subsection{Data Processing and Annotation}
    The collected video data are preprocessed according to the following procedure: (1) since the videos are recorded continuously under all-weather conditions, video segments that do not contain ship targets are first removed; (2) videos captured during berthing that mainly contain non-maritime backgrounds are discarded to avoid introducing data irrelevant to navigation scenarios; (3) the remaining videos are then sampled into frames. To reduce redundancy among images, one frame is retained every 1–5 s for each video; (4) the extracted images are further filtered by removing those without ship targets, with severe blur, or with high visual similarity; (5) to obtain high-quality image samples, Step (4) is repeated until all retained images satisfy the quality requirements; (6) finally, cross-validation is performed to check the consistency and integrity of all images, and all images are uniformly named using seven-digit numbers, e.g., 0000001.jpg.

    \begin{figure*}[!t]
        \centering
        \includegraphics[width=1.0\linewidth]{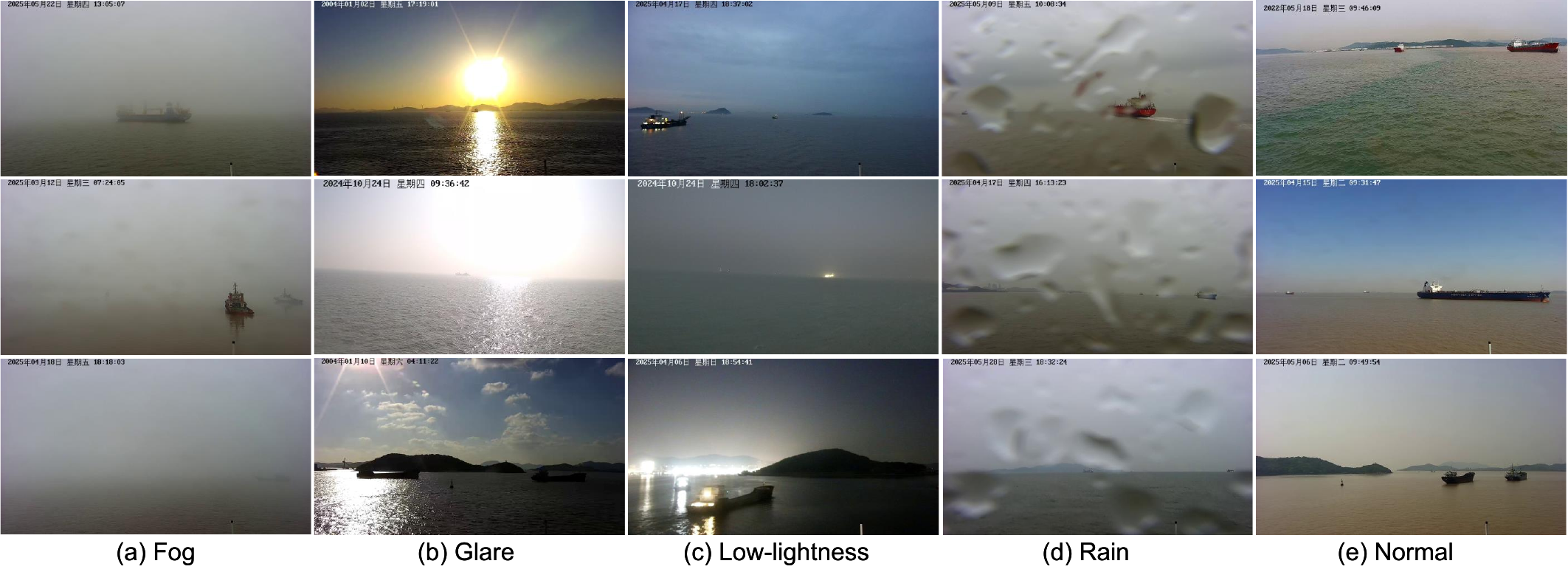}
        \caption{Image examples in WUTDet under fog, glare, low-lightness, rain, and normal conditions. These diverse imaging conditions provide data support for evaluating the performance of different models under heterogeneous visual conditions.}
        \label{fig: different scene}
    \end{figure*}

    With respect to the annotation strategy, all ship targets in WUTDet are labeled as a single category, “ship.” This choice is motivated by two main considerations. On the one hand, the dataset is designed for maritime situational awareness and collision avoidance, in which ship type is not a primary concern. On the other hand, the dataset contains a large number of small-scale targets with areas ranging from 0 to 32² pixels, for which it is difficult to reliably distinguish specific ship categories during annotation. The LabelImg\footnote{https://github.com/HumanSignal/labelImg} tool is used to annotate ship targets with tight rectangular bounding boxes, and the category labels together with their spatial locations are saved in extensible markup language (.xml) files following the same format as the Pascal VOC 2007 dataset, with all “difficult” fields set to 0.

    To ensure the standardization and accuracy of the annotations, a three-level quality control mechanism consisting of manual labeling, cross-validation, and expert review is adopted. The specific rules are as follows: (1) for occluded targets and ship clusters, only the visible parts are annotated, without inferring the locations of the occluded regions; (2) multiple annotators with experience in object detection independently label all images in the first round; (3) multiple rounds of cross-validation are conducted to eliminate missing and incorrect annotations, and the bounding boxes are refined to obtain more accurate results; (4) experts randomly inspect the annotations to evaluate their quality; (5) the final annotation files are named using the same seven-digit numbers as the corresponding images, e.g., 0000001.xml.

    \subsection{Dataset Statistics}
    As illustrated in Fig. \ref{fig: Samples in WUTDet}, all images in WUTDet are collected from real ship navigation processes, covering multiple typical operational states, including port area, anchorage area, berthing state, and navigation state. During the data screening stage, samples are carefully selected to ensure broad coverage of diverse imaging variations, such as different object scales, ship parts, illumination conditions, viewpoints, background complexity, and occlusion, thereby guaranteeing the representativeness and challenge of the dataset in real-world maritime scenarios.

    \textbf{Dataset Split.} To ensure sufficient training data and reliable performance evaluation, the large-scale ship dataset WUTDet contains a total of 100,576 high-resolution images and 381,378 annotated instances. As shown in Table \ref{tab: data split}, the dataset is strictly divided into training, validation, and test sets with a ratio of 8:1:1. Specifically, the training set includes 80,461 images with 305,264 object instances, accounting for 80.0\% of the entire dataset, which provides adequate supervision for deep learning models to capture complex ship characteristics and maritime backgrounds. The validation and test sets contain 10,058 and 10,057 images, with 38,118 and 37,996 instances, respectively. During the splitting process, particular attention is paid to maintaining consistent data distributions across the training, validation, and test sets in terms of object categories, scale distributions, and scene complexity. This strategy effectively avoids evaluation bias caused by data imbalance and ensures that the experimental results are representative and stable.

    \textbf{Scene diversity.} To comprehensively evaluate the generalization ability and robustness of detection models under complex maritime environments, the dataset is further divided into specific scene subsets according to different meteorological conditions and illumination levels. As shown in Table \ref{tab: scene statistics}, while keeping the training set unchanged, samples from the validation and test sets are merged to construct a multi-dimensional evaluation benchmark consisting of fog, glare, low-lightness, rain, and normal scenes. In terms of statistical distribution, the normal scene contains 14,688 images with 59,183 object instances, forming the main basis for evaluation. For more challenging adverse conditions, the dataset includes 3,380 fog images (9,688 objects), 900 glare images (2,995 objects), 694 rain images (2,125 objects), and 578 low-lightness images (2,294 objects). Representative image samples of WUTDet under different scene conditions are shown in Fig. \ref{fig: different scene}. Such fine-grained scene partitioning aims to simulate extreme visual disturbances encountered in real navigation scenarios. It not only quantitatively demonstrates the wide environmental coverage of WUTDet but also provides a solid experimental basis for in-depth analysis of model performance degradation and failure modes under specific conditions.

    \begin{table}[t]
        \centering
        \caption{Image and object statistics under different scene conditions in WUTDet.}
        \label{tab: scene statistics}
        \renewcommand{\arraystretch}{1.5}
        \begin{tabular*}{\linewidth}{@{\extracolsep{\fill}}lccccc}
        \hline
         Scene         & Fog          & Glare          & Low-lightness      & Rain          & Normal          \\
        \hline
        Images         & 3,380        & 900            & 578                & 694           & 14,688          \\
        Objects        & 9,688        & 2,995          & 2,294              & 2,125         & 59,183          \\
        \hline
        \end{tabular*}
    \end{table}

    \begin{figure}[!t]
        \centering
        \includegraphics[width=1.0\linewidth]{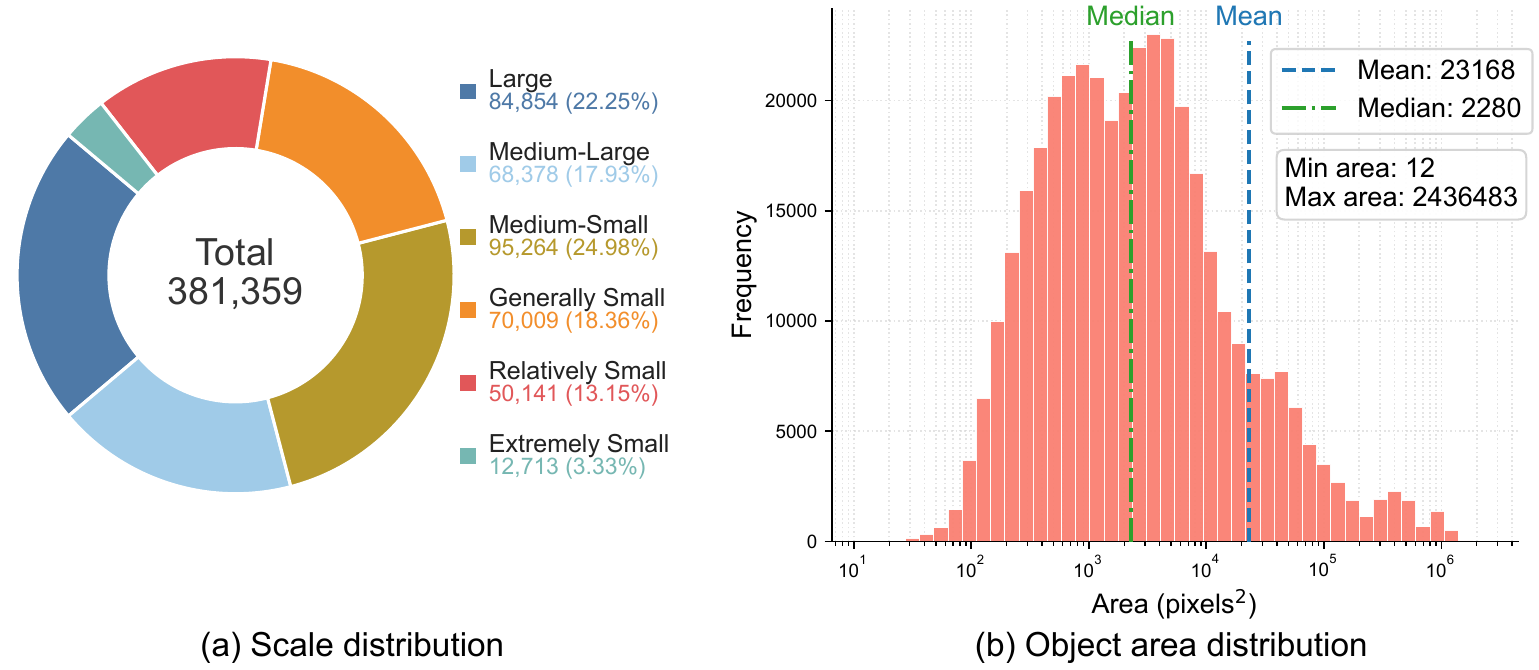}
        \caption{Presents the statistics of object scales in WUTDet. (a) shows the number and proportion of objects in different area intervals, where Extremely Small, Relatively Small, Generally Small, Medium-Small, Medium-Large, and Large correspond to area ranges of $[0, 12^2]$, $(12^2, 20^2]$, $(20^2, 32^2]$, $(32^2, 60^2]$, $(60^2, 96^2]$, and $(96^2, +\infty)$ pixels, respectively. (b) illustrates the frequency distribution of object areas.}
        \label{fig: the statistics of object scales}
    \end{figure}

    \textbf{Ship Size Diversity.} To objectively evaluate the model's capability in representing multi-scale targets, especially its performance in detecting distant and small objects, we conduct a detailed statistical analysis of the object scale distribution in WUTDet. The dataset covers an extremely wide range of target sizes, with instance areas ranging from as small as 12 pixels to as large as 2,436,483 pixels. As shown in Fig. \ref{fig: the statistics of object scales} (a), the objects are categorized into six scale levels. Among them, extremely small, relatively small, and generally small targets together account for 34.81\% of all annotated instances. This high proportion of small-scale targets makes the dataset especially suitable for benchmarking small-object detection algorithms in complex maritime scenarios. Medium-scale targets dominate the dataset, where medium-small and medium-large objects jointly constitute 42.91\% of the total instances, ensuring stable learning and robust evaluation under conventional detection settings. In addition, large targets (22.25\%) further enhance the diversity of object scales. Fig. \ref{fig: the statistics of object scales} (b) presents the frequency histogram of object areas under a logarithmic coordinate system, which illustrates a continuous distribution with an evident Gaussian-like tendency, and a major concentration in the range of $10^3$ to $10^4$ pixels. Overall, while maintaining a relatively balanced scale distribution, WUTDet emphasizes small targets through deliberate sample construction. This characteristic not only challenges existing algorithms in multi-scale representation learning but also provides a high-quality data foundation for accurate recognition of distant vessels in maritime surveillance tasks.

    \begin{figure}[!t]
        \centering
        \includegraphics[width=1.0\linewidth]{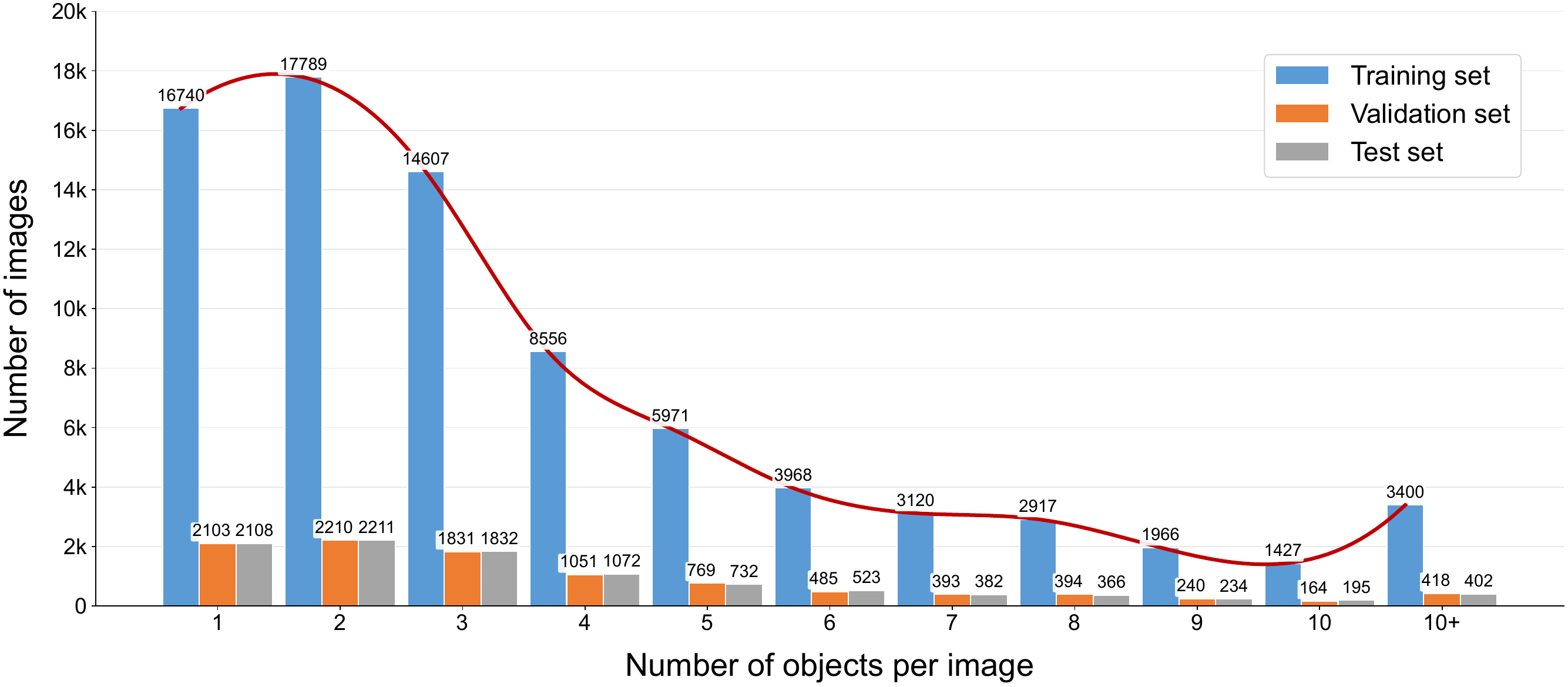}
        \caption{Distribution of the number of objects per image in WUTDet. The bars show the statistics of the training, validation, and test sets, while the red curve denotes a smoothed approximation of the object-count distribution in the training set.}
        \label{fig: number of objects per image}
    \end{figure}

    \begin{figure}[!t]
        \centering
        \includegraphics[width=1.0\linewidth]{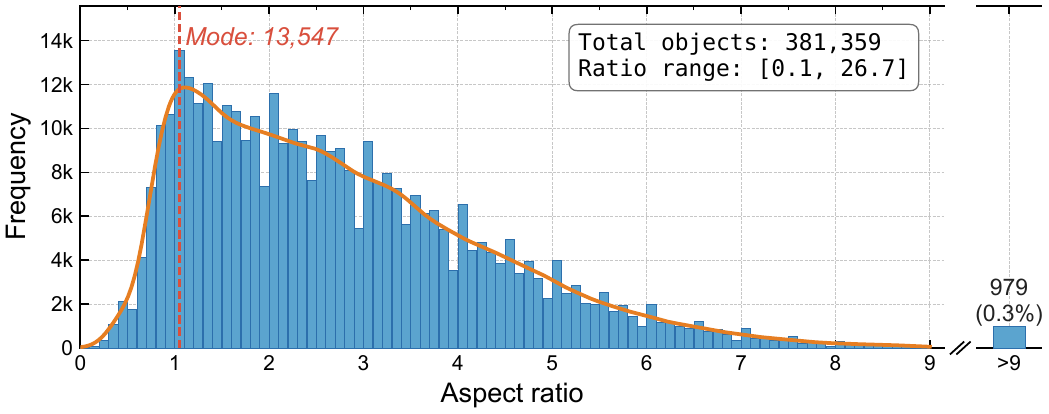}
        \caption{Aspect ratio distribution of objects in WUTDet. The aspect ratio is defined as the height-to-width ratio of the bounding box. The histogram shows the frequency distribution of aspect ratios, and the orange curve indicates the corresponding smoothed trend.}
        \label{fig: object aspect ratio distribution}
    \end{figure}

    \textbf{Object Distribution.} To comprehensively analyze the statistical characteristics of the WUTDet dataset and the challenges it poses for detection tasks, we conduct a quantitative evaluation from two perspectives: object density and geometric distribution. First, the analysis of per-image object density is illustrated in Fig. \ref{fig: number of objects per image}. WUTDet exhibits a wide range of distributions from sparse to highly dense scenes, and the number of objects per image follows a pronounced long-tail pattern. In the training, validation, and test sets, images containing 1–3 objects account for the majority, with the highest frequency observed for images containing two objects (17,789 images in the training set). Notably, the dataset also includes a substantial number of crowded scenes: approximately 3,400 training images and 820 validation/test images contain more than 10 objects (10+). Such high-density object distributions realistically simulate complex maritime traffic conditions in ports and anchorages, where vessels frequently converge and overlap, thereby imposing higher requirements on the algorithm’s robustness to occlusion under complex spatial arrangements. Second, statistical analysis of the geometric properties reveals that the aspect ratio distribution of WUTDet spans an extremely wide dynamic range, from 0.1 to 26.7. As shown in Fig. \ref{fig: object aspect ratio distribution}, although most ship instances are concentrated within the range of $[1, 5]$, a considerable number of annotations exhibit extreme aspect ratios. This distribution reflects significant geometric variations caused by viewpoint changes (drastic transitions between frontal and side views) and by differences in vessel types (slender ocean-going cargo ships versus short and wide working vessels). In summary, the high-density clustering characteristics and extreme geometric variations observed in WUTDet not only enhance the realism and difficulty of the dataset, but also compel detection models to achieve stronger geometric robustness within complex spatial contexts, which is essential for accurate recognition and localization in challenging maritime environments.

\section{Method} \label{sec: method}
\noindent 

    This section reviews three detection paradigms including CNN-based, Transformer-based, and Mamba-based methods, focusing on their architectural evolution, core design principles, and remaining limitations.
    
    \subsection{CNN-Based Methods}
    One-stage convolutional detectors, represented by the YOLO series, unify proposal generation, classification, and bounding-box regression into a single forward pass, formulating detection as end-to-end dense prediction and eliminating the computational overhead of explicit region proposal stages.

    \begin{figure*}
        \centering
        \includegraphics[width=1.0\linewidth]{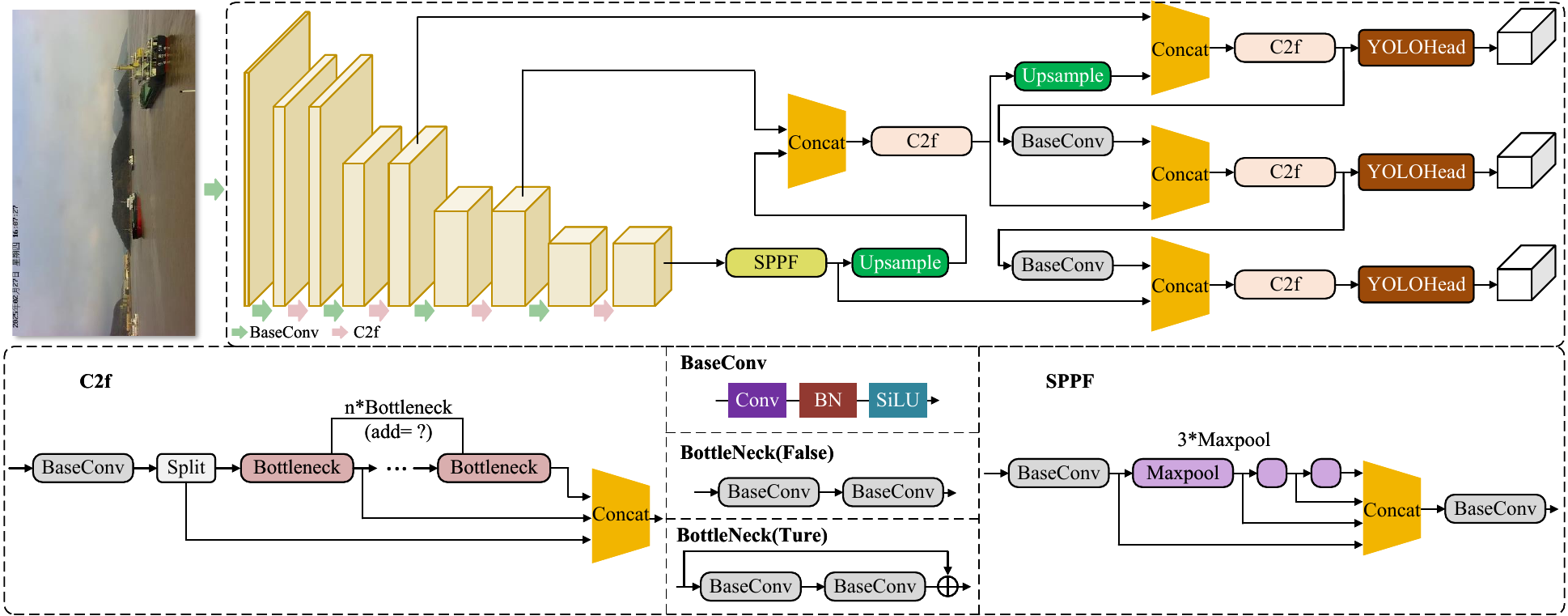}
        \caption{Pipeline of the YOLOv8 architecture \cite{Jocher_Ultralytics_YOLO_2023}. As a representative CNN-based object detection model, YOLOv8 improves detection efficiency and representation capability through designs such as an anchor-free decoupled head and enhanced gradient flow, and has become a widely adopted baseline for real-time object detection tasks.}
        \label{fig: yolov8}
    \end{figure*}

    \textbf{YOLOv8.} YOLOv8 \cite{Jocher_Ultralytics_YOLO_2023} improves upon earlier YOLO variants by addressing two long-standing limitations: insufficient feature reuse under constrained computational budgets and optimization conflicts caused by coupled prediction heads (as illustrated in Fig. \ref{fig: yolov8}). Its backbone incorporates cross-stage partial connectivity to balance representational richness against parameter efficiency, facilitating the interaction between high-level semantic and low-level spatial features, which is particularly critical for detecting small or boundary-ambiguous targets. The neck adopts a bidirectional feature-pyramid topology that propagates complementary cues across resolution levels, sustaining recall under significant scale variation. A decoupled head disentangles classification and localization objectives into separate branches, mitigating gradient conflicts that commonly degrade regression stability in shared-parameter formulations. Nevertheless, as a dense prediction framework, YOLOv8 inherently generates substantial candidate redundancy, necessitating non-maximum suppression (NMS) whose hyperparameter sensitivity can compromise performance in densely packed scenes.

    \textbf{YOLOs.} YOLOX \cite{ge2021yolox} eliminates anchor priors and adopts a decoupled head to reduce inductive bias, pursuing architectural consistency with its end-to-end training objective. YOLOv6 \cite{li2022yolov6} targets the accuracy-latency trade-off through hardware-aware operator design and refined feature reuse in the backbone and neck stages. YOLOv11 \cite{Jocher_Ultralytics_YOLO_2023} restructures the backbone module topology and feature propagation paths to improve gradient flow and representational capacity within a comparable computational envelope. Hyper-YOLO \cite{feng2024hyper} extends the standard YOLO feature hierarchy with enhanced fusion and context modeling modules, strengthening discriminability under cluttered backgrounds and severe scale disparity. YOLOv12 \cite{tian2025yolov12} and YOLOv13 \cite{lei2025yolov13} jointly optimize backbone and fusion architectures with a focus on multi-scale semantic coherence and computational redundancy reduction under lightweight deployment constraints. FBRT-YOLO \cite{xiao2025fbrt} refines the regression pathway through dedicated feature enhancement modules, improving bounding-box stability for hard-to-localize samples. YOLO-MS \cite{chen2025yolo} augments cross-scale fusion pathways with explicit multi-resolution branches, enhancing robustness to target size variation.

    \subsection{Transformer-Based Methods}
    Transformer-based methods, represented by the DETR \cite{carion2020end} family, reformulate detection as set prediction, where a fixed set of learnable queries interact with image features to produce predictions trained end-to-end via bipartite matching. This formulation fundamentally bypasses anchor design and heuristic candidate filtering, but introduces new challenges in convergence efficiency, computational cost on high-resolution features, and small-object sensitivity.

    \begin{figure*}
        \centering
        \includegraphics[width=0.9\linewidth]{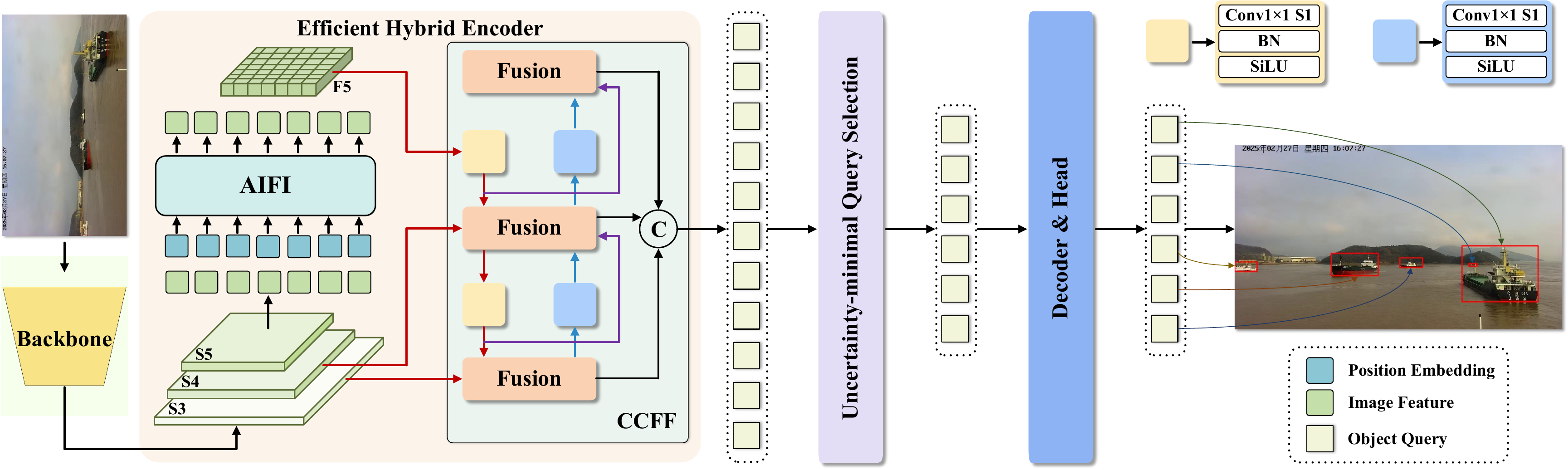}
        \caption{Pipeline of the RT-DETR architecture \cite{zhao2024detrs}. As a representative Transformer-based object detection model, RT-DETR \cite{zhao2024detrs} improves computational efficiency and spatial alignment through designs such as the efficient hybrid encoder and IoU-aware query selection, and achieves end-to-end real-time detection without post-processing by eliminating NMS.}
        \label{fig: rtdetr}
    \end{figure*}

    \textbf{RT-DETR.} RT-DETR \cite{zhao2024detrs} targets the computational bottleneck that prevents prior DETR variants from real-time deployment (as shown in Fig. \ref{fig: rtdetr}). Its core design is an Efficient Hybrid Encoder that explicitly decouples intra-scale feature interaction from cross-scale feature fusion, substantially reducing the quadratic cost of multi-scale Transformer encoding while retaining contextual richness. An IoU-aware query selection mechanism further replaces confidence-based initialization with localization-quality-guided priors, improving spatial alignment between initial queries and target regions. By preserving the set prediction paradigm, RT-DETR eliminates heuristic post-processing such as NMS entirely, and additionally supports flexible speed-accuracy control by adjusting decoder depth at inference without retraining. However, the aggressive efficiency constraints impose a representational ceiling. In severely crowded scenes or under extreme small-object density, the streamlined encoder may compromise multi-scale feature alignment relative to heavier non-real-time counterparts, exposing a sensitivity to the balance between encoder capacity and feature resolution.

    \textbf{DETRs.} Deformable DETR \cite{zhu2020deformable} addresses the slow convergence and prohibitive complexity of the original DETR by replacing dense global attention with deformable attention that sparsely aggregates features from a small set of learnable sampling points across multiple scales. DAB-DETR \cite{liu2022dab} reformulates query representation by explicitly associating each query with a spatial reference point, which conditions attention aggregation and positional updates in the decoder, thereby strengthening spatial correspondence between queries and target regions. DINO \cite{zhang2022dino} extends multi-scale modeling by feeding four-scale features into both encoder and decoder and stabilizing query-feature interactions, yielding more reliable matching and regression under small-object and crowded conditions. D-FINE \cite{peng2024d} and its related variants target localization quality specifically, introducing iterative box refinement on top of DETR-style predictions to reduce residual regression errors. LW-DETR \cite{chen2024lw} prioritizes deployment efficiency by coupling a lightweight backbone with lower-complexity interaction modules, serving as a representative baseline for resource-constrained Transformer-based detection. DEIM \cite{huang2025deim} builds upon the D-FINE refinement framework, reinforcing query interaction mechanisms and matching stability to improve prediction consistency and localization robustness.

    \begin{figure*}
        \centering
        \includegraphics[width=0.9\linewidth]{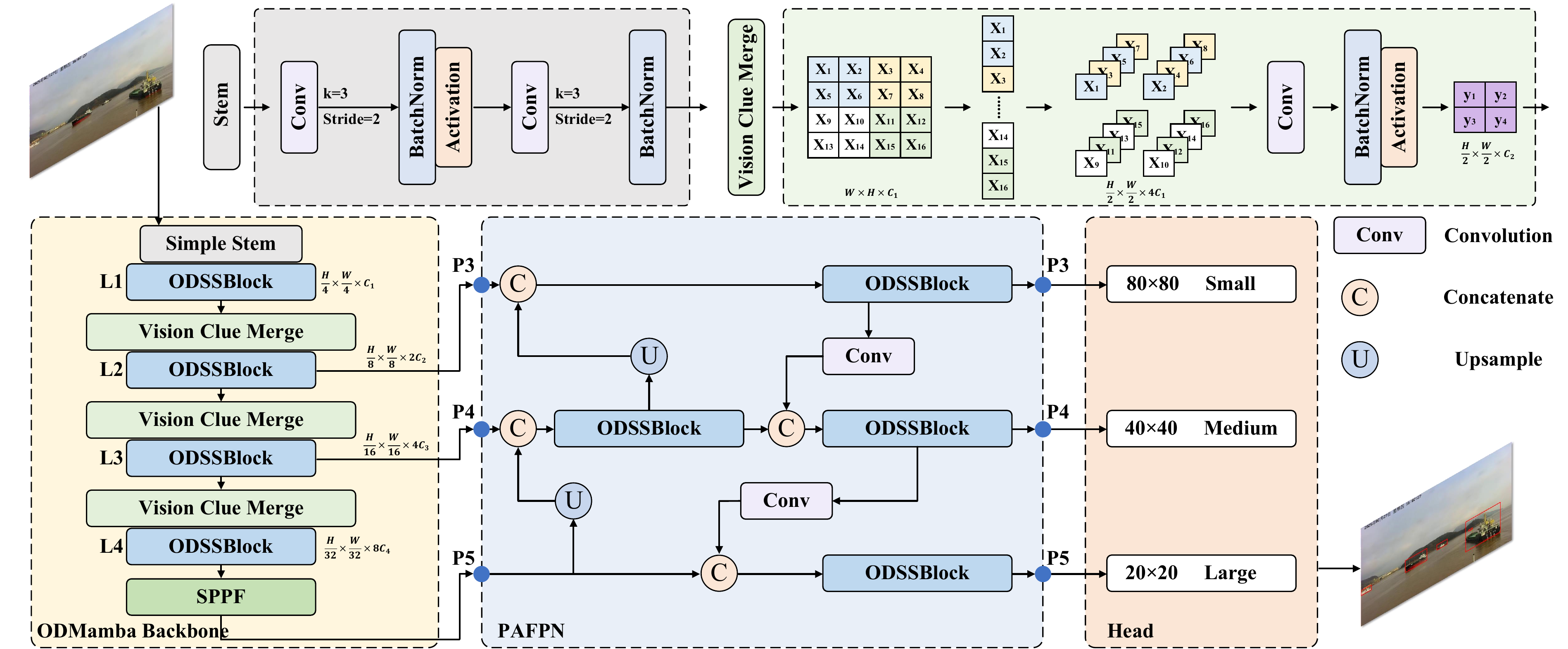}
        \caption{Pipeline of the Mamba-YOLO architecture \cite{wang2025mamba}. As a representative Mamba-based object detection model, Mamba-YOLO \cite{wang2025mamba} introduces SSM to efficiently capture long-range dependencies with linear computational complexity. By avoiding the quadratic overhead of traditional attention mechanisms, it achieves a favorable balance between global contextual modeling and real-time inference efficiency.}
        \label{fig: mambayolo}
    \end{figure*}

    \subsection{Mamba-Based Methods}

    Mamba-based methods introduce selective SSM as an alternative to attention for long-range dependency modeling. Unlike token-wise attention whose computational cost scales quadratically with sequence length, Mamba propagates global context through input-dependent state transitions with linear complexity, offering a favorable trade-off between contextual modeling capacity and computational overhead for high-resolution feature extraction.

    \textbf{Mamba-YOLO.} Mamba-YOLO \cite{wang2025mamba} integrates selective state space modeling into the one-stage YOLO detection framework, targeting the fundamental tension between global context aggregation and real-time inference that neither pure CNN nor Transformer-based detectors fully resolve (as shown in Fig. \ref{fig: mambayolo}). Its backbone replaces standard convolutional blocks with SSM-based modules that propagate long-range structural dependencies along sequential state trajectories, extending the effective receptive field beyond what local convolution kernels can achieve without incurring the quadratic cost of self-attention. To mitigate the directional sensitivity inherent in sequential state propagation over two-dimensional feature maps, Mamba-YOLO adopts multi-directional scanning strategies that traverse spatial features along complementary paths, improving the isotropy of contextual coverage. The detection head retains the dense prediction paradigm of YOLO, preserving inference efficiency while benefiting from globally enriched feature representations. However, the sequential state recurrence limits parallelism relative to attention-based counterparts, and the scanning strategy remains sensitive to scene structure, particularly under irregular spatial distributions.

    \textbf{Mambas.} VMamba \cite{liu2024vmamba} establishes a general-purpose vision backbone based on selective state space modeling, introducing the cross-scan mechanism to extend one-dimensional state propagation to two-dimensional feature maps while maintaining linear complexity. The tiny variant VMamba-T is paired with Faster R-CNN \cite{ren2016faster} as a two-stage detection baseline, where the SSM-enriched features benefit both region proposal generation and RoI-level classification. MobileMamba \cite{he2025mobilemamba} targets mobile deployment by designing lightweight state space blocks with reduced channel dimensions and simplified state transitions. Its B1 variant is combined with RetinaNet \cite{lin2017focal} as a one-stage anchor-based baseline, trading the localization refinement of two-stage pipelines for inference efficiency. Together, these two configurations represent complementary integration strategies for Mamba backbones within established detection frameworks.

\section{Experiments} \label{sec: experiments}

    This section evaluates the performance of 20 detection methods on WUTDet, and further validates the effectiveness of WUTDet and the generalization capability of the models using the unified test set Ship-GEN.
    
    \subsection{Experimental Data and Experimental Settings}

    \textbf{Experimental Data.} To comprehensively evaluate the performance of different object detection methods in complex maritime scenarios, extensive comparative experiments are conducted on WUTDet. Furthermore, to validate the effectiveness of WUTDet in cross-dataset scenarios, we construct a unified cross-dataset test set, termed Ship-GEN. Given that a high proportion of small objects is a prominent characteristic of WUTDet, we select WaterScene \cite{yao2024waterscenes} and WSODD \cite{zhou2021image}, which exhibit the highest and second-highest proportions of small objects among the datasets listed in Table \ref{tab: dataset comparison} excluding WUTDet, as the data sources for Ship-GEN. Specifically, Ship-GEN is composed of a portion of the WUTDet test set, a portion of the WaterScene test set, and the entire WSODD test set, resulting in a total of 3,276 images. To ensure fairness and consistency, the annotations in WaterScene and WSODD are unified by removing all bounding boxes associated with non-ship categories, while retaining only ship instances and relabeling them as a single category, “ship.” All other experimental settings remain consistent with the original configurations of WaterScene and WSODD. This cross-dataset evaluation protocol provides a more comprehensive benchmark for assessing the generalization ability and robustness of detection models across different data distributions.

    \textbf{Experimental Settings.} All training and testing experiments were conducted on a unified high-performance deep learning workstation equipped with a single NVIDIA GeForce RTX 4090 GPU with 24 GB of memory. All methods were implemented using the PyTorch framework. Model scales were distinguished by the suffixes “-S”, “-T”, or different backbone architectures. According to their architectural characteristics, the evaluated detection methods were categorized into three groups: CNN-based methods, Transformer-based methods, and Mamba-based methods. For CNN-based methods, the input image resolution was uniformly set to $640 \times 640$. CenterNet \cite{zhou2019objects} and YOLOv6 strictly followed their original configurations and were trained for 140 and 400 epochs, respectively, while the remaining CNN-based detectors were trained for 300 epochs using their default settings. Transformer-based models adopted a unified multi-scale input strategy and were trained according to their original configurations, with training epochs ranging from 36 to 132. Mamba-based models retained their standard single-scale training paradigm with only 12 training epochs and were evaluated at two resolutions, $640 \times 640$ and $1333 \times 800$, to analyze their sensitivity to scale variations. Mamba YOLO \cite{wang2025mamba} followed the YOLO training framework with a fixed input resolution of $640 \times 640$ and a training schedule of 300 epochs.

    \subsection{Evaluation Metrics}
    To provide a comprehensive and objective evaluation of the baseline models and the dataset constructed in this work, we adopt multiple evaluation metrics based on the MS COCO benchmark, including $\text{AR}_{50:95}(\%)$, $\text{AP}_{50:95}(\%)$, $\text{AP}_{s}(\%)$, $\text{AP}_{m}(\%)$, $\text{AP}_{l}(\%)$ and FPS. The detailed definitions are given as follows.

    \textbf{Average Recall and Average Precision.} Precision (P) denotes the proportion of correctly predicted positive samples among all predicted positives, while Recall (R) denotes the proportion of correctly detected positives among all ground-truth positives. In our experiments, $\text{AR}_{50:95}(\%)$ and $\text{AP}_{50:95}(\%)$ are used as the primary evaluation metrics. They correspond to the mean Average Precision and mean Average Recall computed over ten different Intersection over Union (IoU) thresholds ranging from 0.50 to 0.95 with a step size of 0.05. This multi-threshold averaging strategy enables a more rigorous assessment of bounding box localization accuracy.

    \textbf{Multi-scale Object Detection.} To objectively evaluate model performance on targets of different sizes, the overall AP is further subdivided according to the pixel area of target instances. Specifically, $\text{AP}_{s}(\%)$ denotes the average precision for small objects (area $< 32^2$ pixels), $\text{AP}_{m}(\%)$ corresponds to medium-sized objects ($32^2 \leq$ area $\leq 96^2$ pixels), and $\text{AP}_{l}(\%)$ represents the average precision for large objects (area $> 96^2$ pixels).

    \textbf{Model Parameters and Inference Speed.} Model size and inference speed are important indicators for assessing practical applicability. The number of parameters (Params, in M) denotes the total number of trainable parameters in a model and reflects its storage cost and computational complexity. Frames Per Second (FPS) is used to measure real-time inference capability and is generally defined as the number of images processed per unit time. In this work, the batch-based average $FPS$ is adopted as the inference efficiency metric, which is calculated according to eq. (\ref{eq: fps}).
    
    \begin{equation}
        FPS=\frac{1}{n}\sum_{k=1}^{n}\left(\frac{Batch Size}{t_{\text{infer}}^{(k)}}\right)
        \label{eq: fps}
    \end{equation}
    where $Batch Size$ denotes the number of input images in a single inference batch, $t_{\text{infer}}^{(k)}$ denotes the inference time of one batch in the $k$-th trial, and $n$ denotes the number of test runs. For fair comparison, $Batch Size$ and $n$ are fixed to 16 and 100, respectively.

    \subsection{Results and Analysis}
    \textbf{Overall performance of different methods on WUTDet.} To systematically evaluate the performance and robustness of object detection methods with different architectures in complex maritime scenarios, a unified comparison experiment was conducted on WUTDet covering three mainstream paradigms, namely CNN-, Transformer-, and Mamba-based models. The quantitative results are summarized in Table \ref{tab: overall performance}.
    
    From the perspective of $\text{AP}_{50:95}(\%)$, Transformer-based methods achieve the best overall performance. Specifically, DEIM-D-FINE-S \cite{huang2025deim} and D-FINE-S \cite{peng2024d} obtain the highest accuracies of 71.8\% and 71.4\%, respectively, which are significantly superior to those of CNN- and Mamba-based models, indicating their stronger capability in modeling object semantics and structural features under complex maritime backgrounds. Among CNN-based approaches, YOLOX-S \cite{ge2021yolox} (67.6\%) and YOLOv6-S \cite{li2022yolov6} (64.0\%) yield the best performance, whereas earlier models such as CenterNet \cite{zhou2019objects} (54.2\%) exhibit significantly lower accuracy, suggesting that conventional convolutional architectures encounter performance bottlenecks in multi-scale and densely distributed scenarios. Mamba-based models show relatively lower accuracy overall; for instance, VMamba-T-FasterRCNN@1x \cite{liu2024vmamba} achieves only 43.9\% with an input resolution of 640$\times$640, while its performance improves to 60.0\% at 1333$\times$800, demonstrating a strong sensitivity to input resolution.

    In terms of $\text{AR}_{50:95}(\%)$, Transformer-based approaches remain dominant. Both DEIM-D-FINE-S (77.1\%) and D-FINE-S (77.0\%) achieve the highest recall, indicating more stable object coverage under dense ship distributions and occlusion conditions. For CNN-based methods, YOLOX-S (70.9\%) and YOLOv6-S (67.6\%) show relatively high recall, whereas Hyper-YOLO-S (29.4\%) exhibits an abnormally low recall rate, revealing severe missed detections in complex scenes. Mamba-based models show limited recall under low-resolution input, e.g., VMamba-T-FasterRCNN@1x achieves only 46.9\% at 640×640, which increases to 64.1\% at higher resolution, indicating insufficient sensitivity to fine-grained targets.

    Regarding multi-scale detection performance ($\text{AP}_{s}(\%)$, $\text{AP}_{m}(\%)$, $\text{AP}_{l}(\%)$), small-object detection poses greater challenges. Transformer-based models show a clear advantage on $\text{AP}_{s}(\%)$, with DEIM-D-FINE-S reaching 56.7\%, outperforming the best CNN-based method YOLOX-S (50.0\%) by approximately 6.7\%, which verifies the effectiveness of self-attention mechanisms in capturing global context and weak object features. In contrast, Mamba-based models perform poorly on small objects; for example, VMamba-T-FasterRCNN@1x achieves only 3.7\% at 640×640, and even with higher resolution its $\text{AP}_{s}(\%)$ increases merely to 34.5\%, remaining significantly lower than those of Transformer and CNN methods. For medium-scale targets ($\text{AP}_{m}(\%)$), the performance gap among different architectures becomes smaller, while Transformer-based methods still maintain an advantage; for instance, DEIM-D-FINE-S (73.1\%) exceeds YOLOX-S (69.8\%). For large targets ($\text{AP}_{l}(\%)$), except for Deformable DETR \cite{zhu2020deformable} (74.9\%) and DAB-DETR-R50 \cite{liu2022dab} (49.6\%), most methods achieve accuracies above 80\%, with DEIM-D-FINE-S and D-FINE-S both reaching 88.4\%, indicating that large vessels in WUTDet are easier to recognize and the inter-model performance gap is relatively limited.

    In terms of the trade-off between accuracy and speed, CNN-based methods exhibit clear advantages. YOLOX-S (60.8 FPS) and Hyper-YOLO-S (66.7 FPS) maintain 63\%–67\% $\text{AP}_{50:95}(\%)$ while achieving real-time or even ultra-real-time inference, making them suitable for online maritime monitoring and embedded deployment. By contrast, Transformer-based methods, although more accurate, generally suffer from lower inference speed. For example, DEIM-D-FINE-S (31.2 FPS) and D-FINE-S (30.8 FPS) are applicable to near-real-time scenarios, whereas DINO-4scale \cite{zhang2022dino} (4.5 FPS) and Deformable DETR (6.2 FPS) fail to meet real-time requirements. Mamba-based models demonstrate a compromise between accuracy and efficiency; for instance, Mamba YOLO-M (15.0 FPS, 63.5\% AP) balances detection performance and speed to a certain extent, indicating their potential for real-time detection tasks.

    \begin{table*}[htbp]
        \centering
        \caption{Performance of different object detection methods on our dataset. Since the authors of YOLOv8-S, YOLOv11-S, Hyper-YOLO-S, YOLOv12-S, YOLOv13-S, and FBRT-YOLO-S do not explicitly specify the module definitions of their backbones, the core structures proposed at the backbone stage in each method are uniformly regarded and labeled as the backbone network in this work. FPS is measured with a batch size of 16. $\uparrow$ indicates that higher values correspond to better performance. Bold indicates the best result, and underlining indicates the second-best result.}
        \label{tab: overall performance}
        \renewcommand{\arraystretch}{1.4}
        \setlength{\tabcolsep}{1.2pt}
        \begin{tabular*}{\textwidth}{@{\extracolsep{\fill}} c l l c c c c c c c c c c}
        \toprule
            \textbf{Category}  & \textbf{Method}                                   & \textbf{Backbone}               & \textbf{Year}  & \textbf{Params(M)}  & \textbf{Epoch}  & \textbf{Input size}  & \textbf{FPS}$\uparrow$  & \textbf{AR$_{50:95}$(\%)}$\uparrow$  & \textbf{AP$_{50:95}$(\%)}$\uparrow$  & \textbf{AP$_s$(\%)}$\uparrow$  & \textbf{AP$_m$(\%)}$\uparrow$  & \textbf{AP$_l$(\%)}$\uparrow$  \\
        \midrule
        
        \multirow{10}{*}{CNN}
            & CenterNet \cite{zhou2019objects}                                     & Resdcn-18                       & 2019           & 14.4                & 140             & $640 \times 640$     & 38.0                    & 61.3                                 & 54.2                                 & 31.1                           & 58.1                           & 80.9                           \\
            & YOLOX-S \cite{ge2021yolox}                                           & Modified CSP                    & 2021           & 8.9                 & 300             & $640 \times 640$     & \underline{60.8}       & 70.9                                 & 67.6                                 & 50.0                           & 69.5                           & 86.0                           \\
            & YOLOv6-S \cite{li2022yolov6}                                         & EfficientRep                    & 2022           & 18.5                & 400             & $640 \times 640$     & 33.08                   & 67.6                                 & 64.0                                 & 36.8                           & 69.8                           & 87.2                           \\
            & YOLOv8-S \cite{Jocher_Ultralytics_YOLO_2023}                         & C2f                             & 2023           & 11.1                & 300             & $640 \times 640$     & 60.1                    & 66.2                                 & 62.5                                 & 33.5                           & 69.1                           & 87.2                           \\
            & YOLOv11-S \cite{Jocher_Ultralytics_YOLO_2023}                        & C3K2                            & 2024           & 9.4                 & 300             & $640 \times 640$     & 57.0                    & 66.4                                 & 62.6                                 & 33.9                           & 69.2                           & 87.1                           \\
            & Hyper-YOLO-S \cite{feng2024hyper}                                    & MANet                           & 2024           & 14.8                & 300             & $640 \times 640$     & \textbf{66.7}           & 29.4                                 & 63.1                                 & 34.5                           & 69.6                           & 87.3                           \\
            & YOLOv12-S \cite{tian2025yolov12}                                     & A2C2f                           & 2025           & 9.0                 & 300             & $640 \times 640$     & 30.2                    & 66.1                                 & 62.4                                 & 33.4                           & 68.8                           & 86.9                           \\
            & YOLOv13-S \cite{lei2025yolov13}                                      & DS-C3K2                         & 2025           & 9.0                 & 300             & $640 \times 640$     & 25.6                    & 65.0                                 & 61.3                                 & 32.5                           & 68.0                           & 86.6                           \\ 
            & FBRT-YOLO-S \cite{xiao2025fbrt}                                      & FCM                             & 2025           & 2.9                 & 300             & $640 \times 640$     & 45.6                    & 66.8                                 & 62.9                                 & 34.3                           & 69.3                           & 87.1                           \\
            & YOLO-MS-S \cite{chen2025yolo}                                        & Modified CSP                    & 2025           & 8.7                 & 300             & $640 \times 640$     & 18.4                    & 64.7                                 & 58.1                                 & 27.6                           & 64.7                           & 85.1                           \\
        \midrule
        
        \multirow{7}{*}{Transformer}
            & \makecell[l]{Deformable\\ DETR \cite{zhu2020deformable}}             & ResNet50                        & 2021           & 39.8                & 50              & multi-scale          & 6.2                     & 64.8                                 & 56.8                                 & 38.5                           & 60.6                           & 74.9                           \\
            & DAB-DETR-R50 \cite{liu2022dab}                                       & ResNet50                        & 2022           & 43.4                & 50              & multi-scale          & 11.1                    & 39.4                                 & 27.6                                 & 6.3                            & 33.8                           & 49.6                           \\
            & DINO-4scale \cite{zhang2022dino}                                     & ResNet50                        & 2023           & 46.6                & 36              & multi-scale          & 4.5                     & 73.3                                 & 65.2                                 & 49.5                           & 67.4                           & 82.6                           \\
            & \makecell[l]{RT-DETR\\-R50VD-M \cite{zhao2024detrs}}                 & PResNet                         & 2024           & 36.4                & 72              & multi-scale          & 10.0                    & 71.3                                 & 65.9                                 & 46.0                           & 68.4                           & 86.0                           \\
            & D-FINE-S \cite{peng2024d}                                            & HGNetv2                         & 2024           & 10.1                & 132             & multi-scale          & 30.8                    & \underline{77.0}                     & \underline{71.4}                     & \underline{56.1}               & \underline{72.8}               & \textbf{88.4}                  \\
            & LW-DETR-S \cite{chen2024lw}                                          & Vit-tiny                        & 2024           & 14.2                & 60              & multi-scale          & 14.8                    & 63.4                                 & 58.7                                 & 35.9                           & 61.9                           & 81.5                           \\
            & DEIM-D-FINE-S \cite{huang2025deim}                                   & HGNetv2                         & 2025           & 10.2                & 132             & multi-scale          & 31.2                    & \textbf{77.1}                        & \textbf{71.8}                        & \textbf{56.7}                  & \textbf{73.1}                  & \textbf{88.4}                  \\
        \midrule
        
        \multirow{5}{*}{Mamba}   
            & \makecell[l]{VMamba-T\\-FasterRcnn@1x \cite{liu2024vmamba}}          & VMamba-T                        & 2024           & 46.7                & 12              & $640 \times 640$     & 7.9                     & 46.9                                 & 43.9                                 & 3.70                           & 56.1                           & 81.1                           \\
            & \makecell[l]{VMamba-T\\-FasterRcnn@1x \cite{liu2024vmamba}}          & VMamba-T                        & 2024           & 46.7                & 12              & $1333 \times 800$    & 3.0                     & 64.1                                 & 60.0                                 & 34.5                           & 66.0                           & 82.1                           \\
            & \makecell[l]{MobileMamba-B1\\-RetinaNet@1x \cite{he2025mobilemamba}} & \makecell[l]{Mobile\\Mamba-B1}  & 2025           & 25.4                & 12              & $640 \times 640$     & 11.8                    & 58.5                                 & 48.2                                 & 21.4                           & 52.1                           & 80.4                           \\
            & \makecell[l]{MobileMamba-B1\\-RetinaNet@1x \cite{he2025mobilemamba}} & \makecell[l]{Mobile\\Mamba-B1}  & 2025           & 25.4                & 12              & $1333 \times 800$    & 6.1                     & 68.3                                 & 62.9                                 & 39.9                           & 65.9                           & 82.5                           \\
            & Mamba YOLO-M \cite{wang2025mamba}                                    & ODMamba                         & 2025           & 21.8                & 300             & $640 \times 640$     & 15.0                    & 67.0                                 & 63.5                                 & 35.4                           & 69.9                           & \underline{87.4}             \\
        \bottomrule
        \end{tabular*}
    \end{table*}
    
    The comparison across different input resolutions shows that higher resolution significantly benefits small-object detection. For example, when increasing the resolution of VMamba-T-FasterRCNN@1x from $640 \times 640$ to $1333 \times 800$, $\text{AP}_{s}(\%)$ rises from 3.7\% to 34.5\%, and $\text{AP}_{50:95}(\%)$ improves by approximately 16.1\%. A similar trend is observed for MobileMamba-B1-RetinaNet@1x \cite{he2025mobilemamba}, whose $\text{AP}_{s}(\%)$ increases from 21.4\% to 39.9\%. These results indicate that high-resolution input effectively alleviates the difficulty of recognizing distant small targets in maritime scenarios, albeit at the cost of substantial inference speed degradation (often halved or more).
    
    From the perspective of architectural design, CNN-based methods rely on local convolutional features and offer high efficiency and deployment friendliness, but are limited in modeling complex spatial relationships and small objects. Transformer-based methods significantly enhance detection accuracy through global modeling, particularly in dense and small-object scenarios, but suffer from high computational complexity and limited real-time capability. Mamba-based methods aim to model long-range dependencies with linear complexity, partially balancing efficiency and representation power; however, their discriminative capability under complex maritime target distributions still requires further improvement.

    \begin{figure*}
        \centering
        \includegraphics[width=0.92\linewidth]{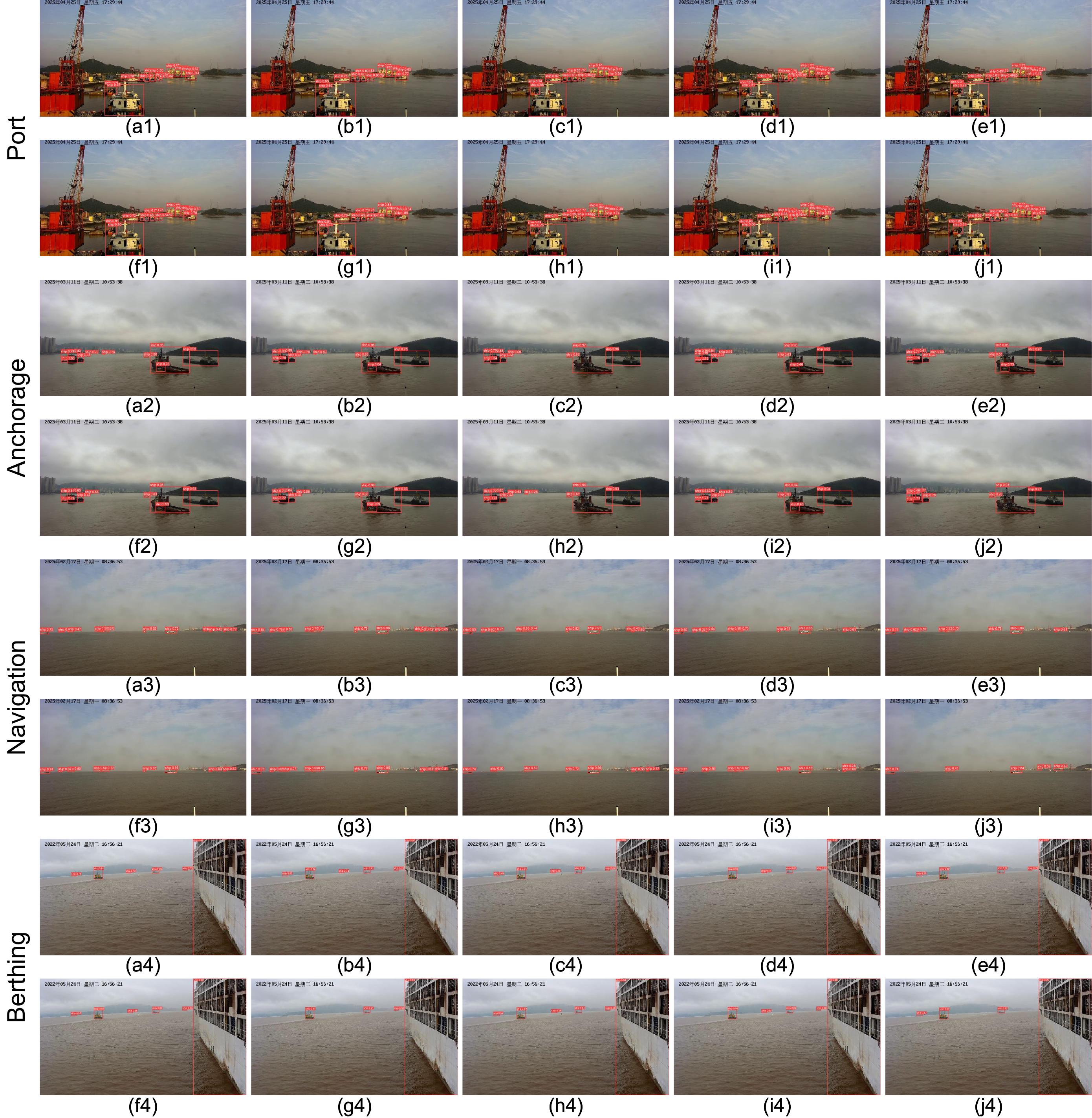}
        \caption{Visualization of detection results of different CNN-based object detection methods on WUTDet. CT denotes the prediction confidence score, whose value is set to the default threshold used for model visualization. The time displayed in the figure does not represent the actual image acquisition time. (a1-a4) represent CenterNet CT = 0.30; (b1-b4) represent YOLOX-S CT = 0.25; (c1-c4) represent YOLOv6-S CT = 0.40; (d1-d4) represent YOLOv8-S CT = 0.25; (e1-e4) represent YOLOv11-S CT = 0.25; (f1-f4) represent Hyper-YOLO-S CT = 0.25; (g1-g4) represent YOLOv12-S CT = 0.25; (h1-h4) represent YOLOv13-S CT = 0.25; (i1-i4) represent FBRT-YOLO-S CT = 0.25; (j1-j4) represent YOLO-MS-S CT = 0.30.}
        \label{fig: visualization 1}
    \end{figure*}

    \begin{figure*}
        \centering
        \includegraphics[width=1.0\linewidth]{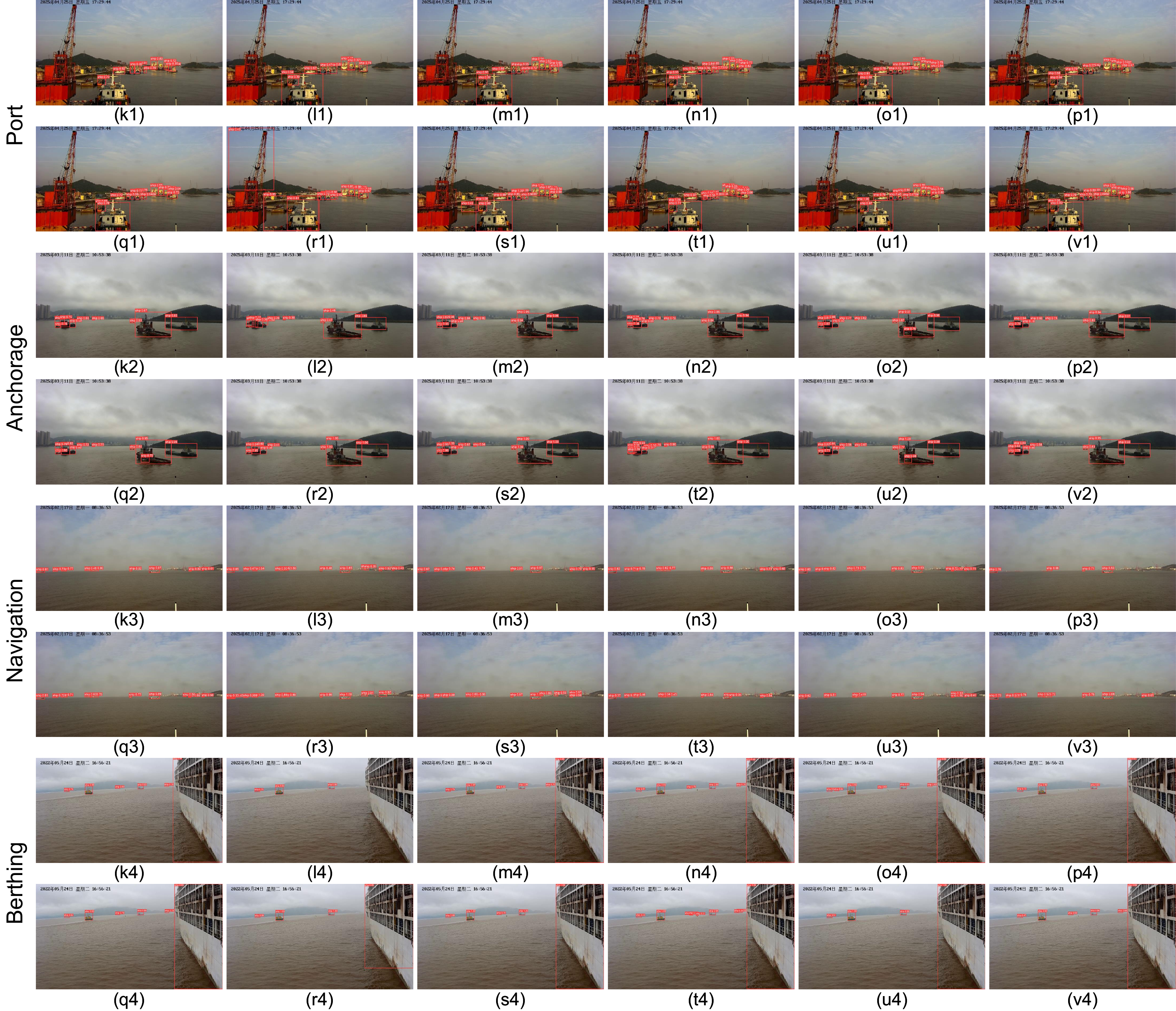}
        \caption{Continuation of Fig. \ref{fig: visualization 1}. Visualization of detection results of different Transformer- and Mamba-based object detection methods on WUTDet. CT denotes the prediction confidence score, whose value is set to the default threshold used for model visualization. (k1-k4) represent Deformable DETR CT = 0.30; (l1-l4) represent DAB-DETR-R50 CT = 0.30; (m1-m4) represent DINO-4scale CT = 0.30; (n1-n4) represent RT-DETR-R50VD-M CT = 0.60; (o1-o4) represent D-FINE-S CT = 0.40; (p1-p4) represent LW-DETR-S CT = 0.50; (q1-q4) represent DEIM-D-FINE-S CT = 0.50; (r1-r4) represent VMamba-T-FasterRcnn@1x CT = 0.30; (s1-s4) represent VMamba-T-FasterRcnn@1x CT = 0.30; (t1-t4) represent MobileMamba-B1-RetinaNet@1x CT = 0.30; (u1-u4) represent MobileMamba-B1-RetinaNet@1x CT = 0.30; (v1-v4) represent Mamba YOLO-M CT = 0.25.}
        \label{fig: visualization 2}
    \end{figure*}

    Figs. \ref{fig: visualization 1} and \ref{fig: visualization 2} present the visualization results of different detection methods on WUTDet. From top to bottom, the images correspond to four typical maritime scenarios, namely anchorage, berth, port, and sailing, respectively, and each scene contains multiple ship targets. Overall, most detection algorithms can effectively recognize ship targets in these scenes; however, certain false positives and missed detections still occur under complex environments. Among the evaluated methods, DEIM-D-FINE-S demonstrates the most stable detection performance across different scenarios. From the perspective of different architectural paradigms, CNN-based detectors show relatively strong capability in handling overlapping or densely distributed ships, but tend to miss small and distant targets. In contrast, Transformer-based detectors generally outperform CNN-based methods in small-object detection, yet exhibit a higher miss rate in ship-overlapping scenarios. Mamba-based detectors show relatively weaker performance in both cases, with more missed detections for overlapping ships as well as distant small targets. In addition, it is worth noting that Deformable DETR and DAB-DETR-R50 also produce relatively more missed detections in some complex scenes, indicating that the robustness of current detection models under densely distributed ships and complicated maritime backgrounds still requires further improvement.

    \textbf{Performance of different methods under various scene conditions on WUTDet.} To further investigate the adaptability and robustness of different detection methods under complex meteorological and illumination conditions, we conduct systematic evaluations on five representative sub-scenarios of WUTDet, including fog, glare, low-lightness, rain, and normal scenes. The quantitative results are reported in Table \ref{tab: different scenarios}.

    \begin{table*}[!htbp]
        \centering
        \caption{Performance of different object detection methods on our dataset under different scenarios. $\uparrow$ indicates that higher values correspond to better performance. Bold indicates the best result, and underlining indicates the second-best result.}
         \label{tab: different scenarios}
        \renewcommand{\arraystretch}{1.4}
        \setlength{\tabcolsep}{2pt}
        \begin{tabular*}{\textwidth}{@{\extracolsep{\fill}} l l  c c c c c c}
        \toprule
            \multirow{2}{*}{Category}     & \multirow{2}{*}{Method}     & \multirow{2}{*}{Input size}     & \multicolumn{5}{c}{AP$_{50:95}$(\%)$\uparrow$}                                                                                                                             \\
                                                                                                          \cmidrule(lr){4-8}                                             
                                          &                             &                                 & Fog                           & Glare                           & Low-lightness                        & Rain                           & Normal                           \\
            \midrule
            \multirow{10}{*}{CNN}
                & CenterNet \cite{zhou2019objects}                      & $640 \times 640$                & 58.2                          & 53.7                            & 61.7                                 & 45.4                           & 53.6                             \\
                & YOLOX-S \cite{ge2021yolox}                            & $640 \times 640$                & 69.9                          & 66.6                            & 69.8                                 & 60.5                           & 67.3                             \\
                & YOLOv6-S \cite{li2022yolov6}                          & $640 \times 640$                & 68.3                          & 65.3                            & \underline{70.4}                     & 58.0                           & 63.1                             \\
                & YOLOv8-S \cite{Jocher_Ultralytics_YOLO_2023}          & $640 \times 640$                & 67.1                          & 63.9                            & 69.9                                 & 56.7                           & 61.7                             \\
                & YOLOv11-S \cite{Jocher_Ultralytics_YOLO_2023}         & $640 \times 640$                & 67.3                          & 63.9                            & 69.9                                 & 56.7                           & 61.8                             \\
                & Hyper-YOLO-S \cite{feng2024hyper}                     & $640 \times 640$                & 67.3                          & 64.0                            & 70.0                                 & 57.2                           & 62.4                             \\
                & YOLOv12-S \cite{tian2025yolov12}                      & $640 \times 640$                & 67.1                          & 63.8                            & 69.7                                 & 56.4                           & 61.5                             \\
                & YOLOv13-S \cite{lei2025yolov13}                       & $640 \times 640$                & 65.9                          & 62.4                            & 68.8                                 & 55.1                           & 60.4                             \\
                & FBRT-YOLO-S \cite{xiao2025fbrt}                       & $640 \times 640$                & 67.1                          & 64.4                            & 70.1                                 & 56.3                           & 62.2                             \\
                & YOLO-MS-S \cite{chen2025yolo}                         & $640 \times 640$                & 62.6                          & 59.3                            & 67.3                                 & 49.8                           & 57.3                             \\
            \midrule
            \multirow{7}{*}{Transformer}
                & Deformable DETR \cite{zhu2020deformable}              & multi-scale                     & 61.2                          & 56.5                            & 61.8                                 & 47.0                           & 56.3                             \\
                & DAB-DETR-R50 \cite{liu2022dab}                        & multi-scale                     & 31.0                          & 29.0                            & 35.9                                 & 19.3                           & 27.1                             \\
                & DINO-4scale \cite{zhang2022dino}                      & multi-scale                     & 66.9                          & 65.2                            & 67.5                                 & 58.0                           & 65.2                             \\
                & RT-DETR-R50VD-M \cite{zhao2024detrs}                  & multi-scale                     & 68.5                          & 65.5                            & 69.8                                 & 58.8                           & 65.6                             \\
                & D-FINE-S \cite{peng2024d}                             & multi-scale                     & \underline{72.4}              & \underline{70.3}                & \textbf{72.9}                        & \underline{65.3}               & \underline{71.5}                 \\
                & LW-DETR-S \cite{chen2024lw}                           & multi-scale                     & 63.1                          & 59.3                            & 63.8                                 & 49.2                           & 58.1                             \\
                & DEIM-D-FINE-S \cite{huang2025deim}                    & multi-scale                     & \textbf{72.9}                 & \textbf{70.7}                   & \textbf{72.9}                        & \textbf{65.5}                  & \textbf{71.9}                   \\
            \midrule
            \multirow{5}{*}{Mamba}
                & VMamba-T-FasterRCNN@1x \cite{liu2024vmamba}           & $640 \times 640$                & 48.2                          & 45.8                            & 61.0                                 & 34.1                           & 42.8                             \\
                & VMamba-T-FasterRCNN@1x \cite{liu2024vmamba}           & $1333 \times 800$               & 64.8                          & 60.8                            & 66.6                                 & 51.1                           & 59.3                             \\
                & MobileMamba-B1-RetinaNet@1x \cite{he2025mobilemamba}  & $640 \times 640$                & 55.2                          & 48.4                            & 61.7                                 & 37.7                           & 46.9                             \\
                & MobileMamba-B1-RetinaNet@1x \cite{he2025mobilemamba}  & $1333 \times 800$               & 65.8                          & 62.6                            & 68.3                                 & 53.5                           & 62.6                             \\
                & Mamba YOLO-M \cite{wang2025mamba}                     & $640 \times 640$                & 67.9                          & 64.8                            & \underline{70.4}                     & 57.9                           & 62.8                             \\
        \bottomrule
        \end{tabular*}
    \end{table*}

    From an overall perspective, Transformer-based methods achieve the best or second-best performance across all scenarios. In particular, DEIM-D-FINE-S attains $\text{AP}_{50:95}(\%)$ values of 72.9\%, 70.7\%, 72.9\%, 65.5\%, and 71.9\% under fog, glare, low-lightness, rain, and normal conditions, respectively, exhibiting the most stable performance among all evaluated methods. This indicates that its globally modeled feature representations are more resilient to complex environmental disturbances. D-FINE-S also maintains consistently high accuracy across all scenarios, further validating the superior cross-scenario generalization capability of Transformer-based detectors.

    Under fog conditions, all models experience performance degradation to varying degrees; however, Transformer-based methods show relatively smaller declines. For example, DEIM-D-FINE-S still achieves 72.9\%, whereas the best-performing CNN-based method, YOLOX-S, reaches 69.9\%. Mamba-based models exhibit clear degradation at low resolution: VMamba-T-FasterRCNN@1x with 640$\times$640 input only achieves 48.2\%, while its performance increases to 64.8\% at 1333$\times$800 resolution, indicating that higher-resolution inputs can partially compensate for the loss of edge information caused by fog-induced blurring.

    Under glare conditions, strong specular reflections significantly interfere with feature discrimination. Among CNN-based methods, YOLOX-S and YOLOv6-S achieve 66.6\% and 65.3\%, respectively, showing relatively stable behavior. Transformer-based methods still outperform others, with DEIM-D-FINE-S reaching 70.7\%. In contrast, DAB-DETR-R50 only achieves 29.0\% in this scenario, which is substantially lower than other Transformer-based methods, suggesting considerable robustness differences among Transformer variants when handling strong reflection interference. 

    Under low-lightness conditions, the proportion of weak-texture and small targets increases, placing higher demands on feature representation capability. The advantage of Transformer-based architectures becomes more pronounced, with both DEIM-D-FINE-S and D-FINE-S achieving 72.9\%, clearly outperforming CNN-based methods such as YOLOX-S at 69.8\%. Mamba-based methods show stronger dependence on input resolution; for instance, MobileMamba-B1-RetinaNet@1x improves from 61.7\% to 68.3\% when the resolution is increased, indicating that fine-grained detail modeling still relies heavily on high-resolution inputs.

    Under rain conditions, performance degradation is observed across all methods due to raindrop occlusion and increased background noise. Transformer-based methods maintain a relative advantage, with DEIM-D-FINE-S achieving 65.5\%, which is notably higher than the best CNN-based result of YOLOX-S at 60.5\%. Mamba-based models show limited improvement in this scenario; for example, VMamba-T-FasterRCNN@1x only reaches 51.1\% at high resolution, revealing insufficient discriminative capability under severe occlusion.

    Under normal conditions, all methods achieve relatively higher performance. Transformer-based detectors remain dominant, with DEIM-D-FINE-S and D-FINE-S achieving 71.9\% and 71.5\%, respectively. The best CNN-based method, YOLOX-S, attains 67.3\%, while Mamba YOLO-M achieves 62.8\%. This indicates that, in the absence of strong environmental interference, the performance gap among different architectures narrows, although Transformer-based models still retain a consistent advantage.

    \begin{figure*}
        \centering
        \includegraphics[width=0.90\linewidth]{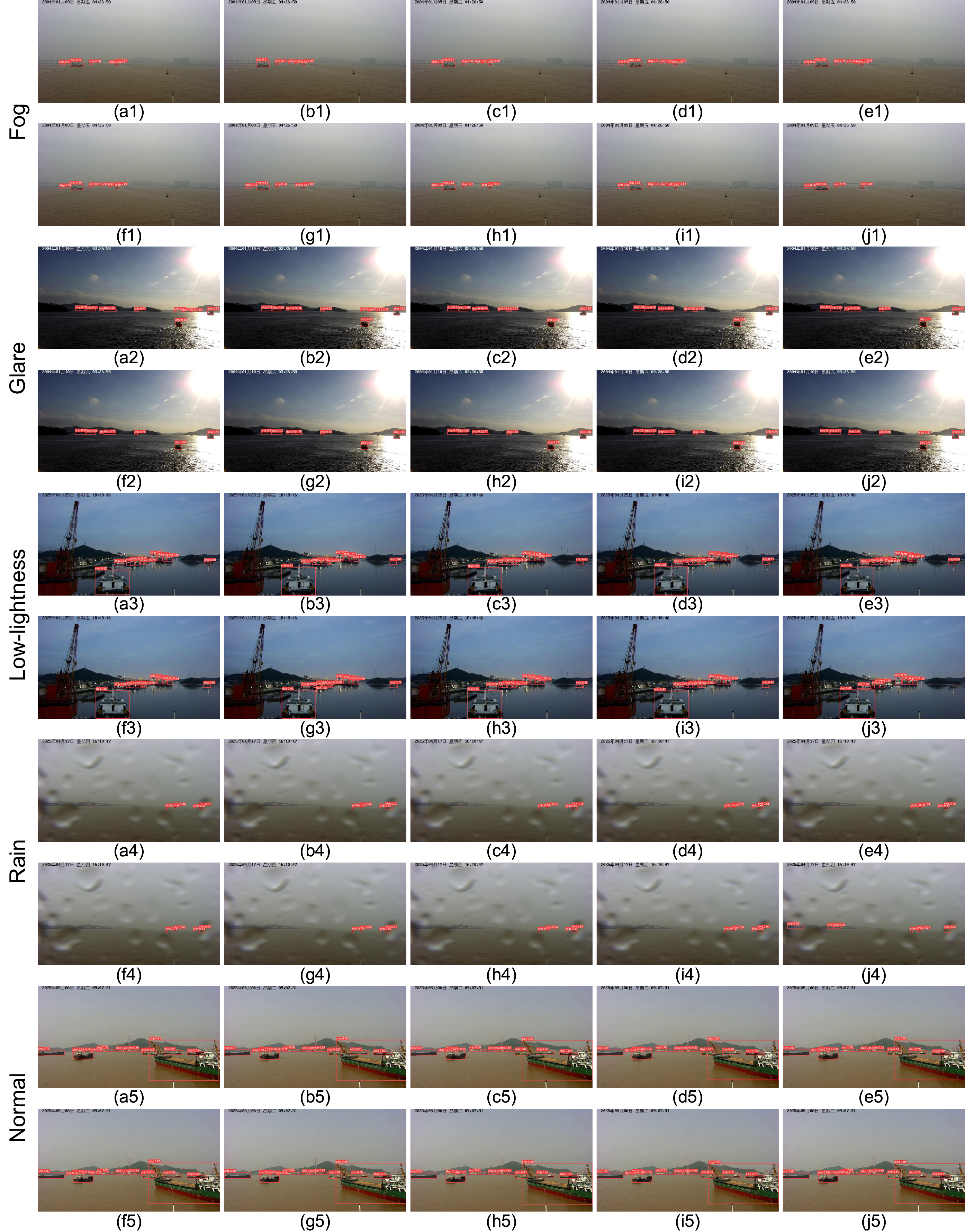}
        \caption{Visualization of detection results of different CNN-based object detection methods on WUTDet under various weather and illumination conditions. CT denotes the prediction confidence score, whose value is set to the default threshold used for model visualization. (a1-a5) represent CenterNet CT = 0.30; (b1-b5) represent YOLOX-S CT = 0.25; (c1-c5) represent YOLOv6-S CT = 0.40; (d1-d5) represent YOLOv8-S CT = 0.25; (e1-e5) represent YOLOv11-S CT = 0.25; (f1-f5) represent Hyper-YOLO-S CT = 0.25; (g1-g5) represent YOLOv12-S CT = 0.25; (h1-h5) represent YOLOv13-S CT = 0.25; (i1-i5) represent FBRT-YOLO-S CT = 0.25; (j1-j5) represent YOLO-MS-S CT = 0.30.}
        \label{fig: visualization 3}
    \end{figure*}

     \begin{figure*}
        \centering
        \includegraphics[width=1.0\linewidth]{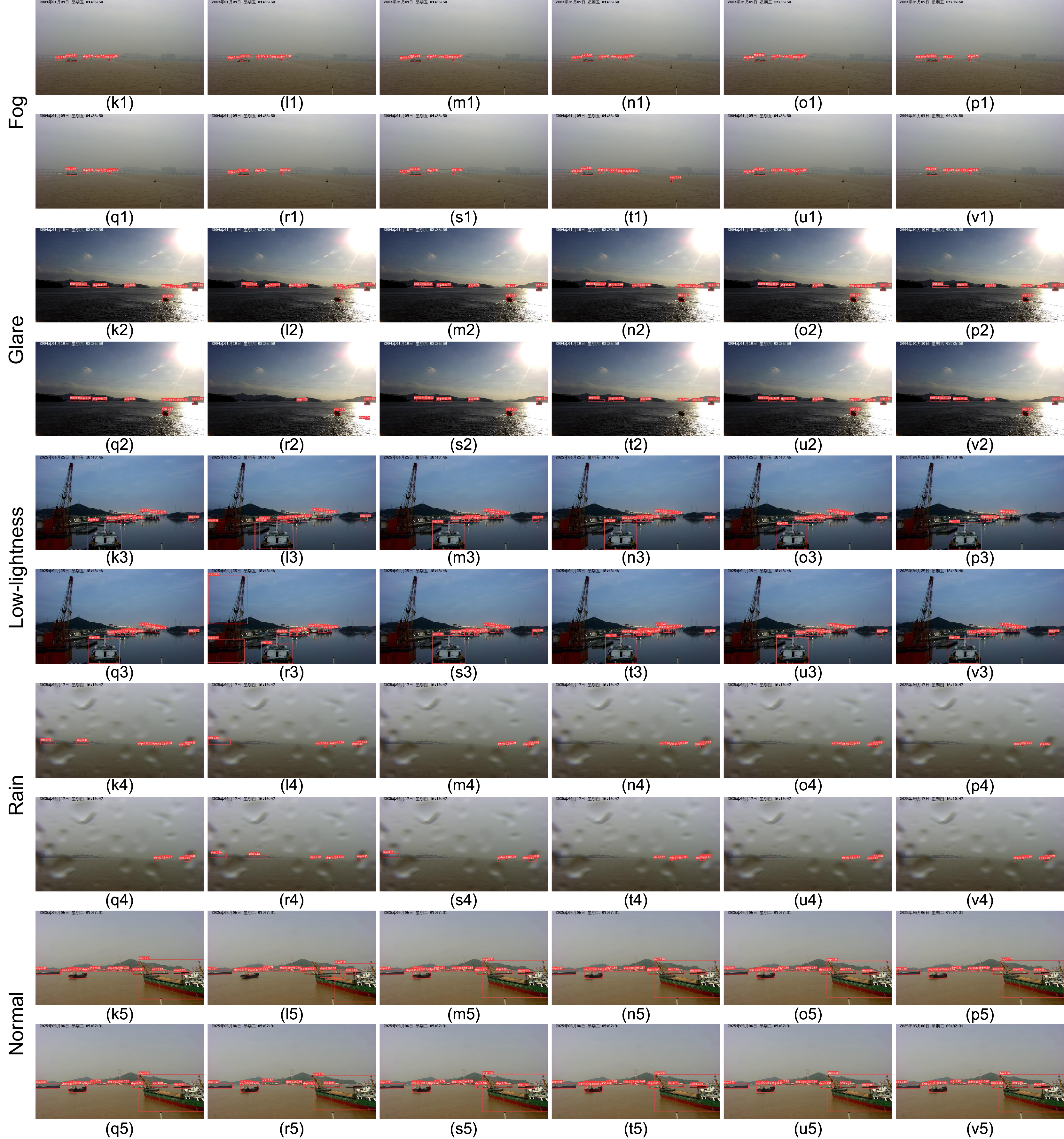}
        \caption{Continuation of Fig. \ref{fig: visualization 3}. Visualization of detection results of different Transformer- and Mamba-based object detection methods on WUTDet under various weather and illumination conditions. CT denotes the prediction confidence score, whose value is set to the default threshold used for model visualization. (k1-k5) represent Deformable DETR CT = 0.30; (l1-l5) represent DAB-DETR-R50 CT = 0.30; (m1-m5) represent DINO-4scale CT = 0.30; (n1-n5) represent RT-DETR-R50VD-M CT = 0.60; (o1-o5) represent D-FINE-S CT = 0.40; (p1-p5) represent LW-DETR-S CT = 0.50; (q1-q5) represent DEIM-D-FINE-S CT = 0.50; (r1-r5) represent VMamba-T-FasterRcnn@1x CT = 0.30; (s1-s5) represent VMamba-T-FasterRcnn@1x CT = 0.30; (t1-t5) represent MobileMamba-B1-RetinaNet@1x CT = 0.30; (u1-u5) represent MobileMamba-B1-RetinaNet@1x CT = 0.30; (v1-v5) represent Mamba YOLO-M CT = 0.25.}
        \label{fig: visualization 4}
    \end{figure*}

    Overall, the multi-scenario experimental results demonstrate that Transformer-based architectures exhibit the strongest stability and generalization capability under diverse meteorological and illumination conditions. CNN-based methods provide a favorable balance between robustness and inference efficiency, whereas Mamba-based detectors still suffer from noticeable performance bottlenecks under adverse environments. These findings further verify the effectiveness and challenge of WUTDet for evaluating model robustness in multi-scenario maritime environments.

    \begin{table*}[htbp]
        \centering
        \caption{Performance of different object detection methods trained on different datasets and evaluated on the Ship-GEN dataset.Note that the input size of MobileMamba-B1-RetinaNet@1x is set to 640$\times$640. $\uparrow$ indicates that higher values correspond to better performance. Purple highlights results obtained on WUTDet, and bold indicates the best result.}
        \label{tab: different datasets}
        \renewcommand{\arraystretch}{1.4}
        \setlength{\tabcolsep}{8.0pt}
        \begin{tabular*}{\textwidth}{l l  c c c c c c}
        \toprule
            Method                                           & Training Set    & Testing Set      & AR$_{50:95}$(\%)$\uparrow$      & AP$_{50:95}$(\%)$\uparrow$      & AP$_S$(\%)$\uparrow$      & AP$_M$(\%)$\uparrow$      & AP$_L$(\%)$\uparrow$                              \\
            \midrule
            \multicolumn{7}{l}{\textbf{Category: CNN}}                                                                                                                                                                                                                            \\
            \midrule
            \multirow{3}{*}{Hyper-YOLO-S \cite{feng2024hyper}}                        & WSODD          & Ship-GEN          & 34.7                            & 26.9                            & 3.5                       & 23.5                      & \textbf{45.1}            \\
                                                                                      & WaterScence    & Ship-GEN          & 29.7                            & 23.6                            & 6.8                       & 17.0                      & 41.0                     \\                                                             
            \rowcolor{lightpurple}                                                    & WUTDet         & Ship-GEN          & \textbf{46.2}                   & \textbf{38.8}                   & \textbf{25.9}             & \textbf{43.0}             & 41.3                     \\
            \multirow{3}{*}{YOLOv6-S \cite{li2022yolov6}}                             & WSODD          & Ship-GEN          & 34.9                            & 25.8                            & 3.4                       & 23.4                      & 42.3                     \\
                                                                                      & WaterScence    & Ship-GEN          & 33.6                            & 25.3                            & 7.3                       & 18.2                      & \textbf{44.1}            \\
            \rowcolor{lightpurple}                                                    & WUTDet         & Ship-GEN          & \textbf{45.7}                   & \textbf{38.1}                   & \textbf{27.6}             & \textbf{42.0}             & 39.8                     \\
            \multirow{3}{*}{YOLOX-S \cite{ge2021yolox}}                               & WSODD          & Ship-GEN          & 36.1                            & 28.7                            & 4.4                       & 25.5                      & \textbf{46.8}            \\
                                                                                      & WaterScence    & Ship-GEN          & 28.0                            & 22.2                            & 8.4                       & 14.0                      & 40.3                     \\
            \rowcolor{lightpurple}                                                    & WUTDet         & Ship-GEN          & \textbf{49.2}                   & \textbf{42.2}                   & \textbf{39.9}             & \textbf{43.0}             & 43.5                     \\
            \midrule
            \multicolumn{7}{l}{\textbf{Category: Transformer}}                                                                                                                                                                                                                    \\
            \midrule
            \multirow{3}{*}{RT-DETR-R50VD-M \cite{zhao2024detrs}}                     & WSODD          & Ship-GEN          & 50.8                            & 36.3                            & 7.5                       & 35.1                      & \textbf{55.3}            \\
                                                                                      & WaterScence    & Ship-GEN          & 42.8                            & 28.8                            & 12.9                      & 19.7                      & 47.4                     \\
            \rowcolor{lightpurple}                                                    & WUTDet         & Ship-GEN          & \textbf{58.3}                   & \textbf{45.2}                   & \textbf{36.2}             & \textbf{45.1}             & 51.1                     \\
            \multirow{3}{*}{D-FINE-S \cite{xiao2025fbrt}}                             & WSODD          & Ship-GEN          & 50.0                            & 32.0                            & 5.5                       & 30.8                      & \textbf{50.3}            \\
                                                                                      & WaterScence    & Ship-GEN          & 41.0                            & 25.3                            & 10.3                      & 17.0                      & 43.7                     \\
            \rowcolor{lightpurple}                                                    & WUTDet         & Ship-GEN          & \textbf{60.7}                   & \textbf{45.3}                   & \textbf{42.6}             & \textbf{44.9}             & 48.2                     \\
            \multirow{3}{*}{DEIM-D-FINE-S \cite{huang2025deim}}                       & WSODD          & Ship-GEN          & 49.3                            & 31.9                            & 5.4                       & 29.5                      & \textbf{51.3}            \\
                                                                                      & WaterScence    & Ship-GEN          & 41.2                            & 25.4                            & 10.0                      & 16.9                      & 45.1                     \\
            \rowcolor{lightpurple}                                                    & WUTDet         & Ship-GEN          & \textbf{61.2}                   & \textbf{46.3}                   & \textbf{44.6}             & \textbf{45.4}             & 49.6                     \\
            \midrule
                \multicolumn{7}{l}{\textbf{Category: Mamba}}                                                                                                                                                                                                                      \\
            \midrule
            \multirow{3}{*}{MobileMamba-B1-RetinaNet@1x \cite{he2025mobilemamba}}     & WSODD          & Ship-GEN          & 31.2                            & 23.3                            & 2.0                       & 17.7                      & \textbf{41.6}            \\
                                                                                      & WaterScence    & Ship-GEN          & 27.8                            & 20.5                            & 2.6                       & 13.8                      & 38.8                     \\
            \rowcolor{lightpurple}                                                    & WUTDet         & Ship-GEN          & \textbf{39.7}                   & \textbf{28.1}                   & \textbf{14.0}             & \textbf{29.9}             & 36.4                     \\
            \multirow{3}{*}{Mamba YOLO-M \cite{wang2025mamba}}                        & WSODD          & Ship-GEN          & 32.1                            & 24.8                            & 3.2                       & 20.4                      & \textbf{42.4}            \\
                                                                                      & WaterScence    & Ship-GEN          & 29.8                            & 23.5                            & 7.9                       & 17.1                      & 39.8                     \\
            \rowcolor{lightpurple}                                                    & WUTDet         & Ship-GEN          & \textbf{44.6}                   & \textbf{37.9}                   & \textbf{25.1}             & \textbf{42.1}             & 40.0                     \\
        \bottomrule
        \end{tabular*}
    \end{table*}

    Figs. \ref{fig: visualization 3} and \ref{fig: visualization 4} present qualitative visualizations of different detection methods on various sub-test sets of WUTDet. From left to right, the scenes correspond to five typical maritime environments, namely fog, glare, low-lightness, rain, and normal conditions. Overall, different environmental factors affect detection performance to varying degrees, among which the Rain scenario has the most significant impact on model performance. In addition, all categories of detection methods exhibit certain levels of missed detections in the Glare scenario, indicating that strong light reflections can significantly interfere with the recognition capability of detection models. Notably, MobileMamba-B1-RetinaNet @ 1× (640$\times$640) demonstrates certain instability across different scenarios. For example, it produces false positives in the Fog scenario, missed detections under Glare conditions, and duplicate detections in the Low Light, Rain, and Normal scenarios.

    \textbf{Overall performance of different methods on Ship-GEN.} To evaluate the generalization ability of different models under cross-distribution conditions, $\text{AP}_{50:95}(\%)$ is adopted as the primary evaluation metric. Eight representative methods are selected from Table \ref{tab: overall performance} for comparison, including CNN-based methods (YOLOX-S, YOLOv6-S and Hyper-YOLO-S), Transformer-based methods (DEIM-D-FINE-S, D-FINE-S and RT-DETR-R50VD-M \cite{zhao2024detrs}), and Mamba-based methods (Mamba YOLO-M and MobileMamba-B1-RetinaNet@1x). The overall performance of these methods on Ship-GEN is reported in Table \ref{tab: different datasets}.

    From the overall accuracy measured by $\text{AP}_{50:95}(\%)$, Transformer-based methods consistently achieve superior performance in cross-dataset scenarios. Specifically, when trained on WUTDet, DEIM-D-FINE-S obtains the highest $\text{AP}_{50:95}(\%)$ of 46.3\%, followed by D-FINE-S (45.3\%) and RT-DETR-R50VD-M (45.2\%), which are markedly higher than those of CNN and Mamba methods. This demonstrates that the Transformer architecture exhibits stronger feature generalization capability and stability when facing distribution shifts. In comparison, the best-performing CNN method, YOLOX-S, achieves 42.2\% after training on WUTDet, slightly inferior to Transformer-based approaches, while YOLOv6-S and Hyper-YOLO-S reach 38.1\% and 38.8\%, respectively, indicating a noticeable performance degradation of convolutional architectures under cross-dataset conditions. Mamba-based models show relatively lower accuracy overall, with Mamba YOLO-M achieving 37.9\% and MobileMamba-B1-RetinaNet@1x only 28.1\%, suggesting that their discriminative capability for cross-dataset detection remains limited.

    In terms of $\text{AR}_{50:95}(\%)$, Transformer-based methods also dominate. DEIM-D-FINE-S and D-FINE-S achieve 61.2\% and 60.7\%, respectively, which are significantly higher than the best CNN method YOLOX-S (49.2\%) and Mamba YOLO-M (44.6\%). This indicates that the Transformer architecture provides more stable target coverage across varying data distributions.

    Regarding scale-specific performance, Ship-GEN poses substantial challenges for small-object detection. For $\text{AP}_{s}(\%)$, Transformer-based methods exhibit clear advantages: DEIM-D-FINE-S and D-FINE-S achieve 44.6\% and 42.6\%, respectively, which are considerably higher than YOLOX-S (39.9\%) and Mamba YOLO-M (25.1\%). This confirms that global modeling mechanisms are more effective for adapting to small-scale ship targets in cross-dataset scenarios. For medium-sized objects ($\text{AP}_{m}(\%)$), Transformer-based methods remain superior, with DEIM-D-FINE-S and D-FINE-S reaching 45.4\% and 44.9\%, exceeding YOLOX-S (43.0\%) and Mamba YOLO-M (42.1\%). For large objects ($\text{AP}_{l}(\%)$), performance gaps among different architectures become smaller; however, Transformer-based methods still retain an advantage, with RT-DETR-R50VD-M achieving 51.1\% and DEIM-D-FINE-S achieving 49.6\%, outperforming CNN and Mamba counterparts.

    From the perspective of training data impact on model generalization, all methods trained on WUTDet achieve substantially better results than those trained on WSODD and WaterScene. For instance, YOLOX-S trained on WUTDet reaches an $\text{AP}_{50:95}(\%)$ of 42.2\%, representing improvements of 13.5 and 20.0\% over WSODD (28.7\%) and WaterScene (22.2\%), respectively. Similarly, DEIM-D-FINE-S attains 46.3\% when trained on WUTDet, compared with 31.9\% on WSODD and 25.4\% on WaterScene. These results indicate that WUTDet provides broader coverage in terms of target scale distribution, scene diversity, and imaging conditions, enabling models to learn more universal and robust discriminative features and thereby significantly enhancing cross-dataset generalization performance.

    Overall, experiments on Ship-GEN demonstrate that Transformer architectures achieve the highest detection accuracy and stability under cross-distribution conditions; CNN-based methods maintain a favorable balance between accuracy and efficiency but exhibit weaker generalization than Transformers; and Mamba-based architectures still suffer from evident performance bottlenecks in cross-dataset detection tasks. These findings further validate the effectiveness of WUTDet in improving model generalization and highlight the value of Ship-GEN as a unified benchmark for evaluating model robustness under cross-dataset settings.

\section{Conclusion} \label{sec: conclusion}
\noindent 

    To address the key limitations of existing ship detection datasets, including restricted scale and a low proportion of small-scale targets, this paper constructs and publicly releases a large-scale, high-resolution ship detection dataset, termed WUTDet. The dataset contains 100,576 images and 381,378 annotated ship instances, covering diverse meteorological conditions and operational scenarios, and thus exhibits substantial complexity and challenge. Based on WUTDet, we conduct a systematic evaluation of 20 baseline models from three mainstream detection paradigms, namely CNN-, Transformer-, and Mamba-based architectures. Experimental results show that Transformer-based models achieve the best performance in both overall detection accuracy and small-object detection, whereas CNN-based models retain a clear advantage in inference efficiency. The emerging Mamba-based architectures, in turn, demonstrate promising potential in balancing detection accuracy and computational efficiency. In addition, cross-dataset experiments based on Ship-GEN further indicate that models trained on WUTDet exhibit stronger generalization capability under different data distributions, thereby validating the effectiveness of the proposed dataset in improving the reliability of ship visual perception systems.

    Although this work establishes a dataset benchmark and conducts a systematic evaluation of representative detection models, several directions remain worthy of further investigation: 1) enriching the ship category annotations in WUTDet to support more fine-grained detection and recognition tasks; 2) extending WUTDet to additional vision tasks, such as instance segmentation and multi-object tracking, to better support complex perception requirements in maritime environments; and 3) designing more efficient detection models that can further improve the detection performance of small-scale ship targets in WUTDet while maintaining high inference efficiency.

\bibliographystyle{IEEEtran}
\bibliography{main}

\end{document}